\documentclass[letterpaper, 10pt, conference]{ieeeconf}  

\IEEEoverridecommandlockouts                              

\usepackage{url}
\usepackage{cite}
\usepackage{xcolor}
\usepackage{graphicx}
\usepackage{wrapfig}
\usepackage[ruled,vlined]{algorithm2e}
\usepackage{algorithmic}

\usepackage{amsmath}
\usepackage{amssymb}
\usepackage{amsopn}
\usepackage{multirow}
\usepackage{rotating}

\usepackage{color}
\usepackage[normalem]{ulem}
\usepackage{listings}

\usepackage{subfigure}

\usepackage[colorlinks=false,pdfborder={0 1 1},letterpaper=true,pagebackref=true]{hyperref}
\usepackage{indentfirst}
\usepackage{grffile}
\usepackage[utf8]{inputenc}

\usepackage[normalem]{ulem}
\usepackage{breqn}

\title{\LARGE \bf  Learning to Herd Agents Amongst Obstacles: Training Robust Shepherding Behaviors using Deep Reinforcement Learning}

\author{Jixuan Zhi$^{1}$ and Jyh-Ming Lien$^{1}$%
\thanks{$^{1}$Jixuan Zhi and Jyh-Ming Lien are with the Department of Computer Science, George Mason University, 4400, University Drive MSN 4A5, Fairfax, VA 22030 USA,
        {\tt\small {\{jzhi,jmlien\}@gmu.edu}}}%
        }

\begin{document}

\maketitle
\thispagestyle{empty}
\pagestyle{empty}
\begin{abstract}
Robotic shepherding problem considers the control and navigation 
of a group of coherent agents (e.g., a flock of bird or a fleet of drones) through the motion of an external robot, called shepherd. 
Machine learning based methods have successfully solved this problem in an empty environment with no obstacles. 
Rule-based methods, on the other hand, can handle more complex scenarios in which environments are cluttered with obstacles and allow multiple shepherds to work collaboratively. However, these rule-based methods are fragile due to the difficulty in defining a comprehensive set of  rules that can handle all possible cases.
To overcome these limitations, we propose the first known learning-based method that can herd agents amongst obstacles.
By using deep reinforcement learning techniques combined with the probabilistic roadmaps, we train a shepherding model using  noisy but controlled environmental and behavioral parameters. Our experimental results show that the proposed method is robust, namely, it is insensitive to the uncertainties originated from both environmental and behavioral models. Consequently, the proposed method has a higher success rate, shorter completion time and path length than the rule-based behavioral methods have. These advantages are particularly prominent in more challenging scenarios involving more difficult groups and strenuous passages. 
\end{abstract}

\section{Introduction}

The robotic shepherding problem, inspired by sheepdogs and sheep, can be defined as one or more shepherd robots try to guide a swarm of agents from an initial location to a goal location\cite{lien2004shepherding}.
Shepherding is applicable to various fields in the real world, such as using robots to herd animals \cite{evered2014investigation,bat2017shepherding}, civil crowd control~\cite{schubert2007decision}, preventing bird strikes near  airports \cite{gade2015herding,gade2016robotic,paranjape2018robotic}, facilitating communication between the unmanned ground vehicles and unmanned aerial vehicles~\cite{chaimowicz2007aerial}.

Algorithms to address the shepherding problem can be classified into two main categories: rule-based and learning-based algorithms. 
Rule-based algorithms define a set of rules that compute the dynamics of the model as a function of the system status. 
These methods are easy to implement and can work well with a large number of the sheep agents.
It has also been shown that multiple  shepherd agents can work collaboratively to herd a large group effectively \cite{lien2005shepherding}.
However, these rule-based methods are fragile due to the difficulty in defining a comprehensive set of  rules that can handle all possible cases.
Some commonly seen difficulties include turning the coherent group in a different direction or removing those trapped agents from the corners.
On the contrary, learning-based algorithms do not need the knowledge about the model and have been shown to be resilient to imperfect group behavior parameters of the sheep. Unfortunately, these learned models have only been trained and tested in simple blank environments. Navigating the sheep among the obstacles has not been considered in these learning-based methods. 



These limitations are the predominant reasons that shepherding robot cannot robustly herd agents amongst obstacles under uncertain behavioral and environmental models.
In this paper, we study these issues in the framework of  deep reinforcement learning. 

Our main contributions include: (1) a  reinforcement learning framework that  trains the shepherding model in the environments populated with agents and obstacles 
influenced by controlled noise based on a simple design of  a reward function that combines the probabilistic roadmap methods with the  reward function, and (2) 
comprehensive experiments demonstrating the learned model that can control the shepherd's movement with higher probability, shorter completion time and path length in herding the sheep to goal.

The remaining of the paper is organized as follows. In Section~\ref{sec:review}, we briefly review two main types of  shepherding methods and multi-agent reinforcement learning. 
Section~\ref{sec:pre} covers the needed notations and background  in group behavior modeling and deep Q-learning. 
The proposed method is described in detail in Section~\ref{sec:method}. 
The experiments and results are presented in Section~\ref{sec:result}. 


\section{Related Work}
\label{sec:review}

Researchers proposed various abstract models of shepherding behaviors in swarm robotics. For instance, Vo et al.~\cite{vo2009behavior} proposed a behavior-based method  which selects intermediate goals on the roadmap of the workspace medial axis. Harrison et al.~\cite{harrison2010scalable} developed a deformable shape model for shepherding control, they modeled the swarm robots as a deformable shape, in their method, the shepherd continually computes the swarm shape and move to the steering points based on the swarm shape. 
Str{\"o}mbom et al. \cite{strombom2014solving} proposed a heuristic method  and described two behaviors: driving and collecting. Driving means steering the flock to the target, and collecting means steering the outlier sheep  back to the flock. Based on the work of Str{\"o}mbom et al, Fujioka and Hayashi\cite{fujioka2016effective} introduced a new method called V-formation control. In their work, shepherd moves among three positions to control the flock along the arc which computed from the center of the flock.

Most shepherding models employ one shepherd to control the flock. However, there are some researchers developed multiple shepherds to guide the flock \cite{lien2005shepherding,pierson2015bio}. 
Recently, Lee and Kim\cite{lee2017autonomous} developed a method without centralized control. In their work, the shepherds only attempt to steer the nearest sheep to the target region without considering other members in the flock, with such behavior rules, multiple shepherds can have an arc formation to control the flock. 
Unfortunately, these rule-based shepherding behaviors are fragile due to the difficulty in defining a set of comprehensive rules to handle all possible cases.


In recent years,  machine learning approaches have been applied to solve shepherding problems. 
Baumann developed a reinforcement learning method \cite{baumann2015learning} called Growing Neural Gas Q-Learning to herd one sheep using a single shepherd. 
He used abstract states to generalize the neighboring and similar states which share the same behavior. Apply Q-learning\cite{watkins1992q} with such small state space, the algorithm works well with linear computational complexity.
Singh et al. \cite{singh2019modulation} proposed a genetic algorithm to optimize shepherd's influence on the sheep, and they showed that their algorithm is energy efficient.   
Based on the Str{\"o}mbom's model\cite{strombom2014solving}, Gee and Abbass\cite{gee2019transparent} proposed a machine learning method that teaches a neural network shepherding skills. They firstly used a human operator to collect training cases, then two different reward systems were designed to learn the sub-skills via reinforcement learning. 
Nguyen et al.\cite{nguyen2019deep} applied inverse reinforcement learning and deep reinforcement learning in UAV shepherding. 
They trained two different models for two shepherding skills of collecting and driving, then aggregated these two models to herd a swarm of ground vehicles. 

None of these learning methods considered obstacles. 

\begin{figure}[th]
{\includegraphics[width = 0.485\textwidth]{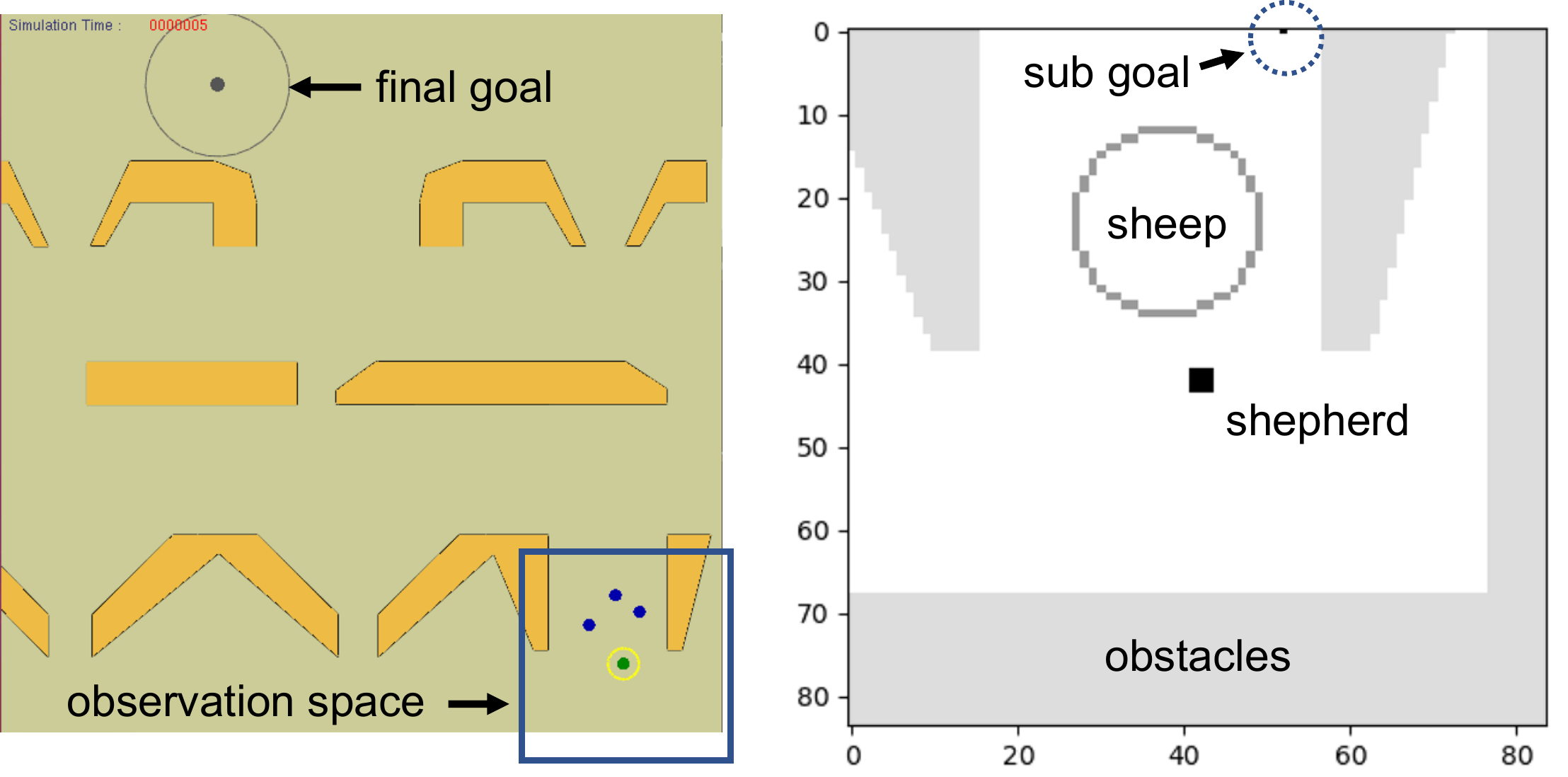}}
\caption{Workspace and observation space.}
\label{fig:ob}
\end{figure}

\section{Preliminaries}
\label{sec:pre}

In our shepherding scenario, there are one shepherd and multiple coherent agents  representing the sheep; and the shepherd has a local view that means shepherd either can know the positions of sheep and goal inside its view or the directions of sheep and goal outside its view. The mission of the shepherd is to guide the sheep to a goal area represented as a circle (See Figure~\ref{fig:ob}).

\subsection{Group Behavior Model}
Reynolds' Boids  \cite{reynolds1987flocks} is used to model the dynamics of the sheep. 
The behavior of each sheep is influenced by the location and velocity of the nearby sheep, which in turn define three control forces: separation, alignment, and cohesion. 
In addition, a \emph{fear} force makes the sheep run away from the shepherd when the shepherd appears in the local neighborhood.
The local neighborhood is defined by the distance and its field of view.
Additional forces are added for obstacle avoidance and damping. 
In each simulation step, these forces are linearly combined and integrated to update the acceleration, velocity and position of each agents.

Changing the coefficients of the weighted sum of these forces  allows us to model different group behaviors. However, tuning these parameter in the Boids model to simulate a particular group of animals (e.g., a flock of birds found in Dulles Airport in Washington DC, USA) is extremely challenging and  often impossible. Therefore, the controller must operate under the assumption that the behavior model of the group is inaccurate. 

\subsection{Deep Q-Learning}
A regular reinforcement learning problem considers four components: (1) a current state $s$ in Observation space, (2) an action $a$ chosen by the agent in Action space, (3) a transition model $P(s'|s,a)$ for next state and (4) a reward $r$ given by the action and state. The optimized policy in each state is found by maximizing the cumulative reward over time. At time $t$, we define the cumulative reward as this quantity:
\[ R_t = r_{t+1} + \gamma r_{t+2}+... = \sum_{k=0}^{\infty} \gamma^k r_{t+k+1}\]
where $\gamma$ is the discount factor.

Deep Reinforcement Learning \cite{mnih2015human} started a new era in reinforcement learning development. Mnih et al. \cite{mnih2015human} use a deep neural network to represent both state and action onto a value, namely $Q(s,a)$. This method can generalize large state spaces that traditional Tabular Q-learning can not. As discussed in by Lien and Pratt \cite{lien2009interactive}, the search space and the dimension of the configuration space for shepherding is extremely large. Therefore, Deep Q-Learning is a necessary tool for solving the shepherding problem. 

Deep Q-Learning uses \emph{replay buffer} of the past experiences and samples training data from it. This method can keep the independent and identically distributed property. Schaul et al.~\cite{schaul2015prioritized} proposed a useful idea to improve the sampling efficiency in the experienced buffer. They proved that by giving priorities to buffer samples based on training loss, they can improve convergence and policy quality.

Deep Q-Learning has two networks: one is the main network, and the other is the target network, which is a copy of the main network and  is used for the Q-value of the next state. The target network is synchronized with the main network every $n$ steps, this method could break the correlation between the steps and make training more stable. Then Van Hasselt et al.~\cite{van2016deep} developed the Double Q-learning technique which can reduce the overestimation of Q-value. They simply use the main network to evaluate action and target network to evaluate value without adding a new network. 
We apply priority replay buffer and double Q-learning techniques in our work.


\section{Method}
\label{sec:method}

\begin{figure}[th]

{\includegraphics[width = 0.45\textwidth]{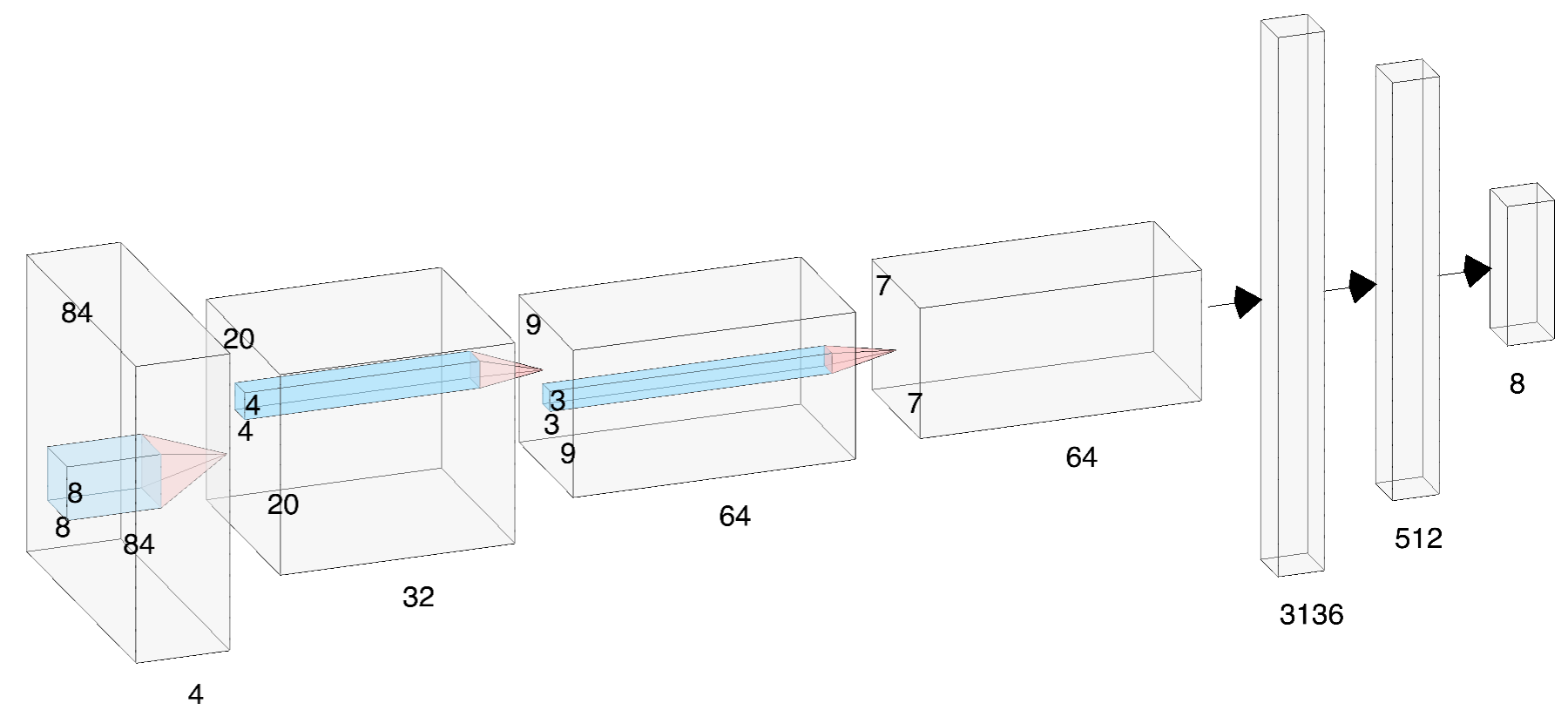}}
\caption{Deep reinforcement learning architecture used in this work.}
\label{fig:architecture}
\end{figure}

\subsection{Model architecture}
Our basic framework follows the model proposed in \cite{mnih2015human}. It is a deep neural network architecture that has three convolution layers and two fully connected layers and there is a rectifier nonlinearity (RELU) unit for each hidden layer. Figure~\ref{fig:architecture} illustrates this model.

\subsection{Observation Space}
\label{subsec:state}
We consider a partially-observable continuous world, in which the shepherd robot can only observe the state of the world in a field of view centered around itself. In practice, we $84\times84$ pixels local frame as input. 
To push our environment back to the Markov Decision Process domain, we use the \emph{frame stack technique} to maintain several frames from the past and use them as the observation at each state. 

\textbf{Representing the Sheep}.
A critical question in training the shepherding behavior is how the group of sheep should be represented. 
We can either view each sheep as a square of $3\times3$ pixels, or aggregate the group, e.g.,  in to a circle.
We speculate that the circle  representation may allow faster convergence in training. 
However, pixels representation may be better for training more precise manipulation needed for difficult environments. 
In terms of the design of reward function, circle representation is also simpler, namely, only  the radius and center are considered, but, for pixels representation, 
we need to calculate the reward for each sheep  and then combine these individual rewards.  
In our experiments, both representations are tested and compared. The circle representation is shown in  Figure~\ref{fig:ob}.

\textbf{Goal and Waypoints}.
In each $84\times84$ frame, we use $3\times3$ pixels in the center of frame to represent the shepherd, one pixel to represent the goal, $3\times3$ pixels to represent a single sheep. If there are multiple sheep, we can use a circle to represent them, the center of a circle is called "sheep center" and the radius is the maximum distance between  sheep and the sheep center. With center and radius, we draw a hollow circle.
In addition, we can choose pixels representation for multiple sheep by $3\times3$ pixels. 
For boundaries and obstacles in the environment, we draw solid polygons. 
In the local frame, we use different pixel values for different objects. See Figure~\ref{fig:ob} for an example. 

The goal (or waypoint) and the sheep are sometime outside of shepherd local field of view, but  these information is essential for the shepherd. 
For the goal, we connect the shepherd position and the goal position as a line segment, and get the intersection of the segment and the boundary of the local view frame, 
then we mark the intersection as the goal inside the local frame. 
For the sheep, if we use circle representation and the circle is partially inside the local frame, we draw that part of the circle in the local frame.
If the circle is completely outside the local frame, we connect the shepherd and sheep center as a line segment, mark the intersection as the sheep in the boundary. 
If we use pixel representation, if some sheep pixels were outside the local frame, we did the same way: connect the shepherd and sheep pixels as line segments, mark the intersection as the sheep pixels in the boundary. 

To handle  environments populated with obstacle, we introduce an available path as instruction: if there is no obstacle between the sheep position and final goal, the path is straight line segment connecting the sheep to the goal; otherwise, the planner will inquire the probabilistic road maps (PRM)~\cite{kavraki1996probabilistic} to find a path between them. We then use the point in this path as the sub-goal to calculate the reward. 

\begin{figure}[t]
\centering
{\includegraphics[width = 0.25\textwidth]{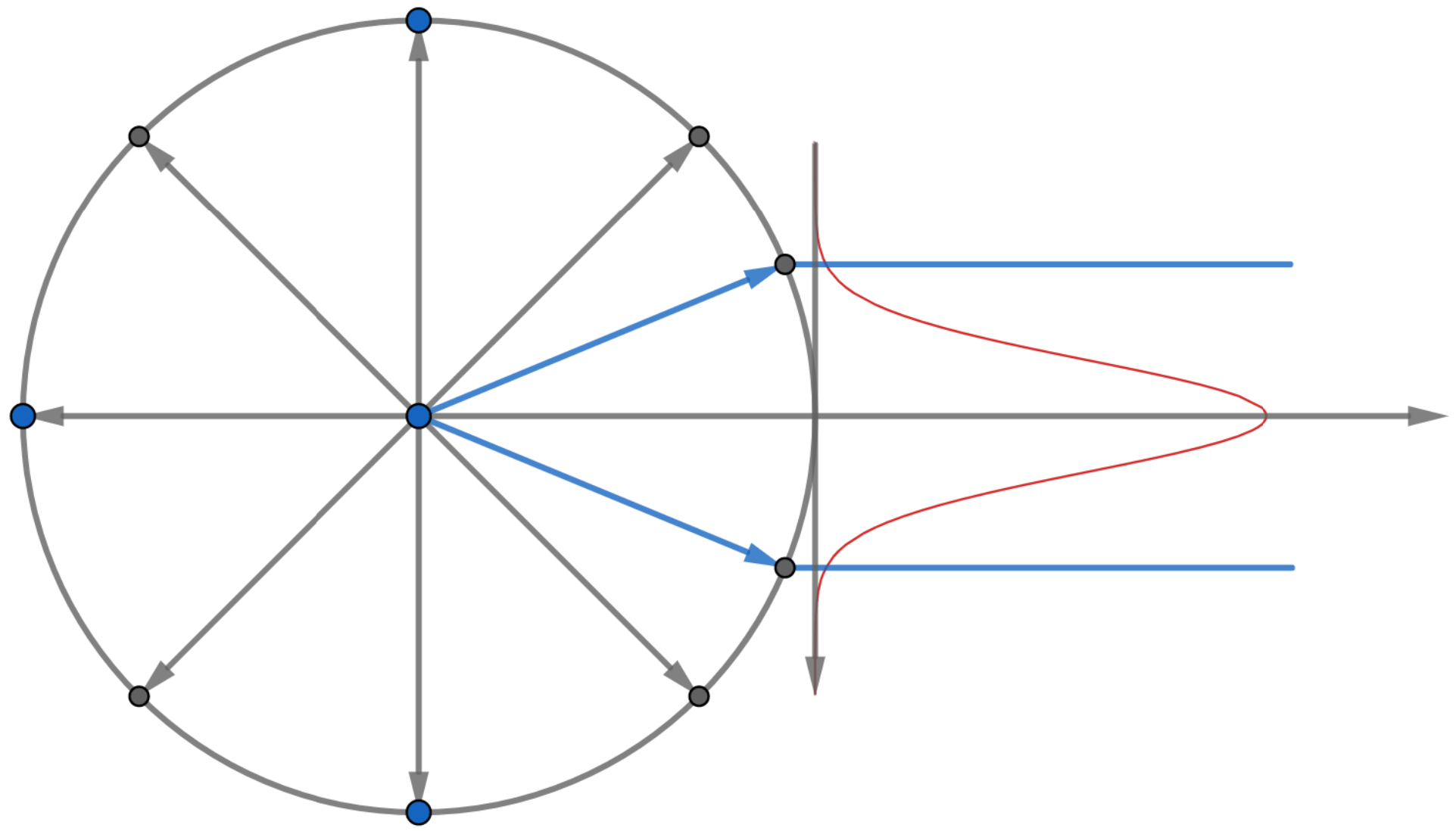}}
\caption{Eight main directions with added noise sampled from a continuous perturbation.}
\label{fig:action}
\end{figure}

\subsection{Action space}

The shepherd robot takes semi-discrete actions as illustrated in Figure~\ref{fig:action}. 
In every step,  one of eight directions is randomly selected and  a  force  along that direction with fixed magnitude is applied to the shepherd. 
In addition, to make the action appear to be continuous, we add a clipped Gaussian noise to the direction of the force.
The deviation of each direction angle is between $-22.5^{\circ}$ and $+22.5^{\circ}$, then we can cover all the entire circle.


\subsection{Reward structure}

The reward function consists of three parts: moving reward, violation reward and goal reaching reward. 
In the following discussion, we mainly focus on reward using the circle representation of sheep, and briefly discuss the rewards in pixel representation. 

In circle representation, we define sheep position for one sheep as its own position, and for multiple sheep as the center of the sheep. 

\subsubsection{Moving Reward}

Moving reward includes two parts: projection reward and path reward.

\begin{figure}[ht]
\centering
{\includegraphics[width = 0.25\textwidth]{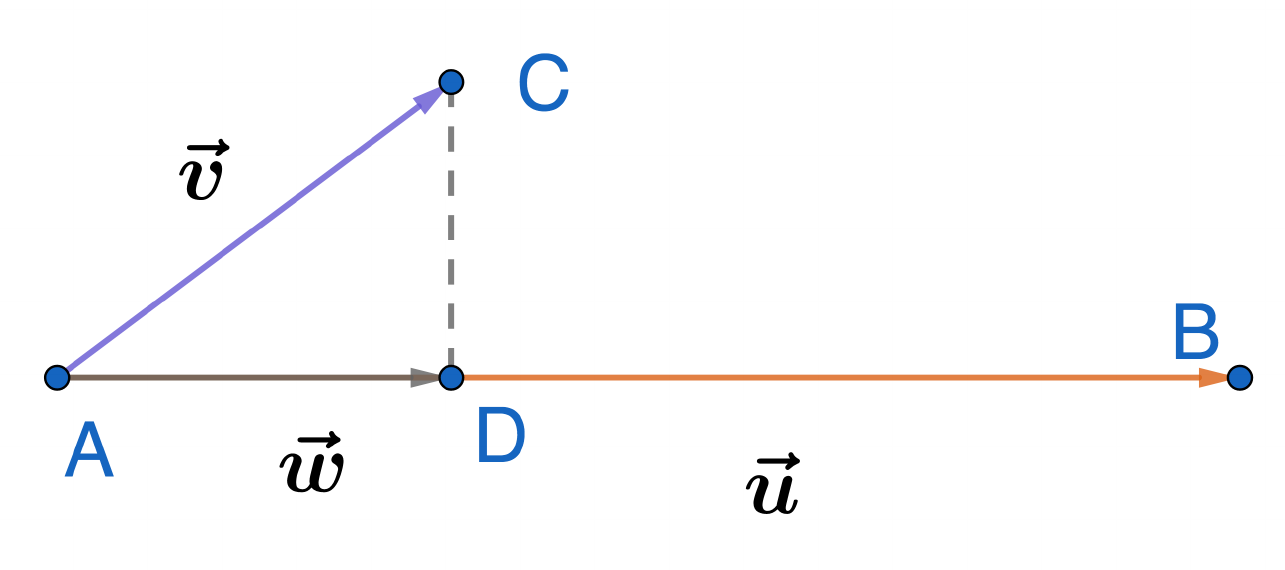}}
\caption{Projection reward}
\label{fig:proj}
\end{figure}

\textbf{Projection Reward}.
To ensure that the sheep is moving in the desired direction and reach the goal as fast as possible, we introduce the projection reward. 
Let the current sheep position be $A$, new sheep position be $C$ and the goal position be $B$. 
We further let  $\vec{u}=B-A$ be the vector from current sheep position to goal position, and let $\vec{v}=C-A$ be the vector form current sheep position to new sheep position.
We compute the projected vector  $\vec{w}$ of vector $\vec{v}$ onto vector $\vec{u}$. See Figure~\ref{fig:proj}. 
We determine the reward based on the direction and length of the vector. 
If the vector $\vec{w}$ has the same direction with vector $\vec{u}$, the reward is positive, otherwise, the reward is negative, i.e., the reward has the same sign as the dot product of 
$\vec{v}$ and $\vec{u}$. 
Note that the negative scale factor (which is $4$ in Alg.~\ref{alg:move}) must be larger than the positive scale factor (which is $2$ in Alg.~\ref{alg:move}) to avoid local optimal. 

\begin{figure*}[ht]
\centering
{\includegraphics[width=0.3\textwidth]{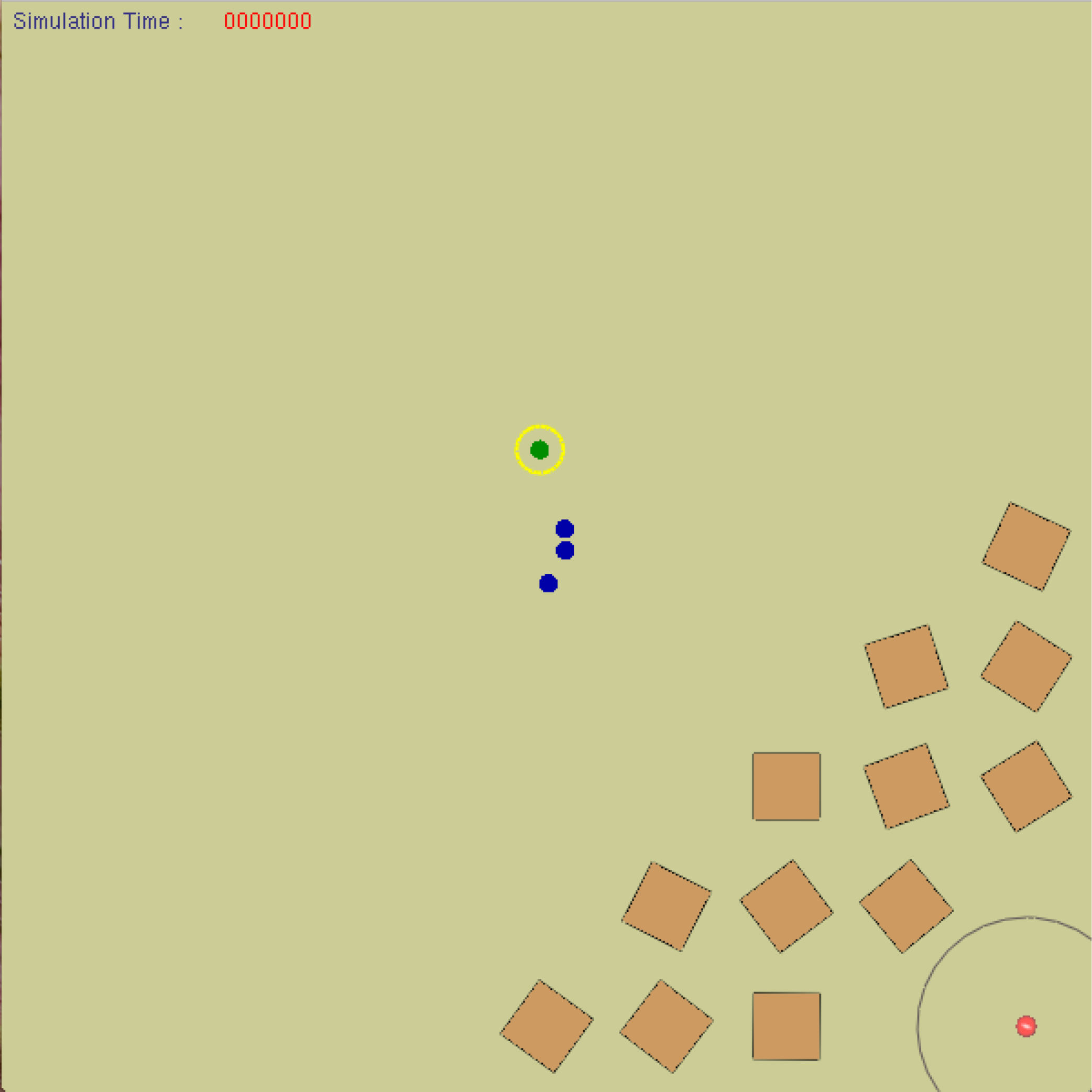}}
{\includegraphics[width=0.3\textwidth]{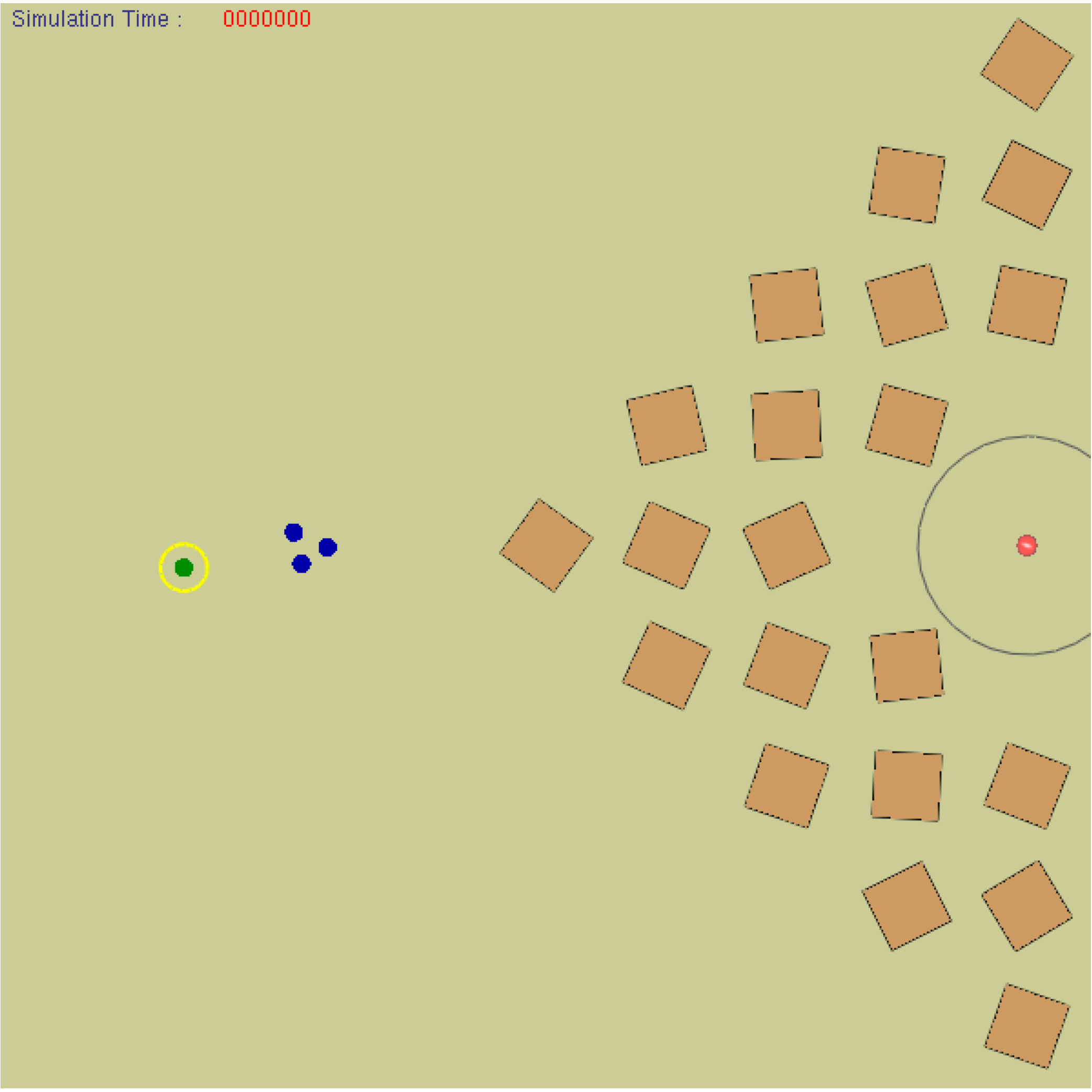}}
{\includegraphics[width=0.3\textwidth]{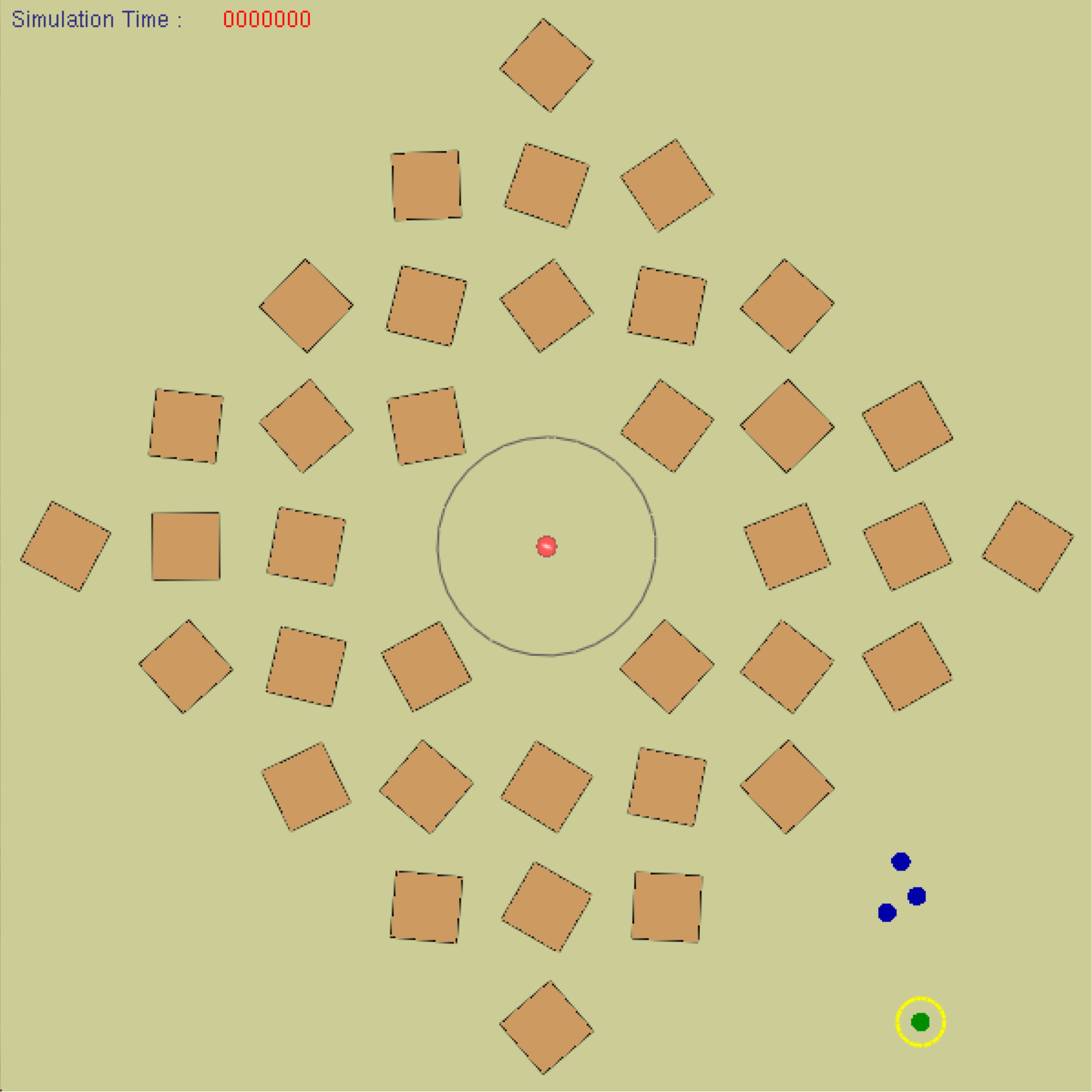}}
\caption{Examples of randomly generated environments,  which contains three layers of obstacles around the randomly sampled goal (from 9 predefined positions) when obstacle density is one. From left to right: corner, edge and center environments.}
\label{fig:random}
\end{figure*}


To encourage exploration in searching for optimal controller, we penalize the action that causes the sheep for staying still or moving too little.
Therefore, we define a small threshold of not moving ($1$ pixel). 

Note that we also define a small threshold of projection length ($1$ pixel). If the projection length is less than the threshold, we give a negative reward, which could avoid the circular motion of the sheep circling around the goal.


\textbf{Path Reward}.
To make the moving reward more accurate, we determine the path reward based on the geodesic distance. The reward function
calculates the difference  of the geodesic distances of two states. 
If the available path of the new state is less than the available path of the old state, the reward is positive, otherwise, the reward is negative. 
We also add a scale factor of the reward and a small threshold ($1$ pixel) to give a small negative reward as same as the projection reward which also can avoid circular motion of the sheep.  See Algorithm~\ref{alg:move} for moving reward of circle representation.

For the pixel representation of sheep, the moving reward only refers to projection reward: for each sheep, we calculate the projection reward as same as that in circle representation, and summarize these awards from individual sheep as the moving reward.

\begin{algorithm}
\SetAlgoLined
\KwIn{ \

a. old \& new positions of sheep center $P^{old}_{Sheep},P^{new}_{Sheep}$ \;
b. old \& new geodesic distances from sheep center to the final goal $D^{old}_{path},D^{new}_{path}$ \;
c. sub goal $P^{Sub}_{Goal}$ \;
d. the minimum move distance threshold $d_{min}$  \;
e. final goal radius $R_{goal}$ \;
}
\KwOut{Moving reward $r_{move}$}
1. vector $\vec{u}=\langle P^{old}_{Sheep}, P^{Sub}_{Goal} \rangle, \vec{v}=\langle P^{old}_{Sheep}, P^{new}_{Sheep}\rangle  $\;
2. let the projection value  $D_{proj} = \vec{v} \cdot \vec{u} / \|\vec{u}\|$ \;
3. let sheep moving difference $D_{diff} = \| \vec{v} \|$ \; 
4. calculate moving reward \;
\begin{algorithmic}
\STATE $r_1=r_2=r_3= 0 $
\IF {$D_{diff}<d_{min}$ }
      \STATE $r_1 =-2 $ 
\ELSE
        \IF {$D_{proj} > d_{min} $}
                \STATE $r_2 = 2 \cdot D_{proj}$
        \ELSIF {$D_{proj} < -d_{min}$}
                \STATE $r_2 = 4 \cdot D_{proj}$
        \ELSE
             \STATE $r_2 = -4 \cdot d_{min} $
        \ENDIF
        
        \IF {$D^{old}_{path}>R_{goal} \  and \  D^{new}_{path} > R_{goal} $}
               \STATE $d_{diff} = D^{old}_{path}-D^{new}_{path} $
                \IF {$d_{diff} > d_{min} $}
                    \STATE $r_3 = 2 \cdot d_{diff}$
                 \ELSIF {$d_{diff} < -d_{min}$}
                    \STATE $r_3 = 4 \cdot d_{diff} $
                \ELSE
                    \STATE $r_3 = -4 \cdot d_{min} $
                 \ENDIF
               
        \ENDIF
\ENDIF
\STATE $r_{move}=r_1+r_2+r_3 $
\end{algorithmic} 
\caption{Moving Reward function}
\label{alg:move}
\end{algorithm}

\subsubsection{Violation Reward and Reaching Goal Reward}

We consider two factors of violation reward. Firstly, for multiple sheep, we need to consider the cohesion of the  group. We define the circle with its center and the radius in subsection~\ref{subsec:state} and penalize the group with a large negative reward if the radius is too large.
This threshold is equal to the view range of the sheep (25 pixels).
Secondly, we want the shepherd to have effective explorations around the group, so we penalize the shepherd with a large negative reward if the shepherd is too far away from the sheep center. The threshold is two times of the view range of the sheep. 

If one of the violation cases occurs or the simulation  exceeds the maximum time step of the training episode, the simulation will terminate this episode and report its failure.

On the contrary, a large positive reward is given when all the sheep reach the goal region and the simulation ends  the episode successfully. 

For the pixel representation, we use the same rules of violation reward and reaching goal reward.

\section{Experiments and Results}
\label{sec:result}
 
In this section, we provide more details on the  training process and highlight the benefits of the proposed method over the rule-based approaches. Our discussion will focus on the effectiveness and robustness of the trained model with random noise injected to our environmental and behavioral models.
All the experiments for training and testing were run on a single desktop, equipped with a NVIDIA TITAN X (Pascal) and an Intel(R) Xeon(R) CPU E5-2630 with 24 logical cores clocked at 2.3Ghz and 32Gb of RAM.

\subsection{Training details}

To  understand how the shepherding controller is trained, let us first discuss how we set the standard parameters used in reinforcement learning. 
We use the ADAM optimizer~\cite{kingma2014adam} with mini batches of size 32, and use $\epsilon$-greedy policy to explore the environments with $\epsilon$ changed linearly from 1.0 to 0.5 for one slope and from 0.5 to 0.02 for the other smaller slope over the first several million frames and fixed at 0.02 thereafter. 
The priority replay buffer size is 200,00, the discount factor is 0.99, and the learning rate is 0.0001. The double Q-learning technique \cite{van2016deep} is also applied.
We use the frame-skipping technique \cite{bellemare2012investigating} to reduce training time. More precisely, the shepherd robot chooses actions on every $k=5$th frame contrary to each frame, and the chosen action is repeated in the intermediate (i.e., skipped) frames. 

For environments with obstacles, we cannot simply connect  the sheep to the final goal and hope that there will be no collision. 
Using the geometry of the shepherd and the workspace, we construct a roadmap that represents the topology of the free space. The roadmap is built using  two variants of probabilistic motion planners that sample one-third of configurations uniformly  \cite{kavraki1996probabilistic} and the other two-thirds on the median axis of the environments \cite{vo2009behavior}. 
In the next section, we will discuss how the nodes in the roadmap can be used to determine the sub-goals during the training phase. 

To increase the robustness of the trained controller, we  consider three types of obstacles in the environment: fixed obstacles, randomly generated obstacles and perturbed obstacles. 
These environments are described below.

\subsubsection{Fixed Obstacles}
We first consider the environments with predefined obstacles. In practice, the environment (such as in a farm or ranch) is usually fixed, and the sheep routinely moves toward one location for grazing and the other location for drinking. Therefore, we create a fixed environment as shown Figure~\ref{fig:ob}. We set two positions and alternate between these two positions as the start and the goal points. 
With this fixed environment, we randomly change the group behavior parameters, namely separation, alignment, and cohesion by the Gaussian distribution and clip method. To be more specific, the range of separation coefficient is between $0.5$ and $1.5$, the range of cohesion coefficient is between $3$ and $7$, the range of alignment coefficient is between $2$ and $4$, and fear coefficient is fixed on $15$. The terminal time steps of each training episode is 6000. The average training time is about 15 hours.

\subsubsection{Randomly Generated Obstacles}
\label{sec:density}
To randomly generate the obstacles in the environments, we divide the environments into a  $9\times 9$ grid. Each grid cell will have a square obstacle or not, and  the orientation of the obstacle is also random, the center of the square obstacle is also randomly generated in a small range between $-2.5$ and $+2.5$ pixels.

For each goal, we randomly create three layers of obstacles around the goal. The number of obstacles is determined by obstacle density. 
Figure~\ref{fig:random} shows the three layers of the obstacle around the goal in three different locations when obstacle density is 1. In our work, we sample the obstacle density from a triangular distribution between 0 and 0.5 with the peak centered at 0.3.

During the training period, we place the goal in 9 different locations of the grids: four corners, four boundary edges, and the center, as illustrated in Figure~\ref{fig:random} . For those 4 corner and 4 boundary edge cases, the initial position of the group is fixed in one grid cell far away from the goal and the obstacles. For the case in which the goal position is at the center, the initial positions of the group could be in one of the four corners.  These initial positions are depicted as  blue points in Figure~\ref{fig:random}.
For every group in these 12 different cells, we position the shepherd in one of the eight directions around the group. Figure~\ref{fig:random} shows the position of shepherd to the south, east and north of group, from left to right figures, respectively. 
For every initial position of  the shepherd or group, we also add a Gaussian noise on these positions in a small range between $-2.5$ and $+2.5$ pixels.

We believe these $12\times 8 = 96$ environments could cover most of the situations in a square environment. The terminal time steps of each training episode is 3000. The average training time is about 20 hours.

\subsubsection{Perturbed Obstacles}
\label{sec:training_perturb}
\begin{figure}[th]
\centering
{\includegraphics[width = 0.23\textwidth]{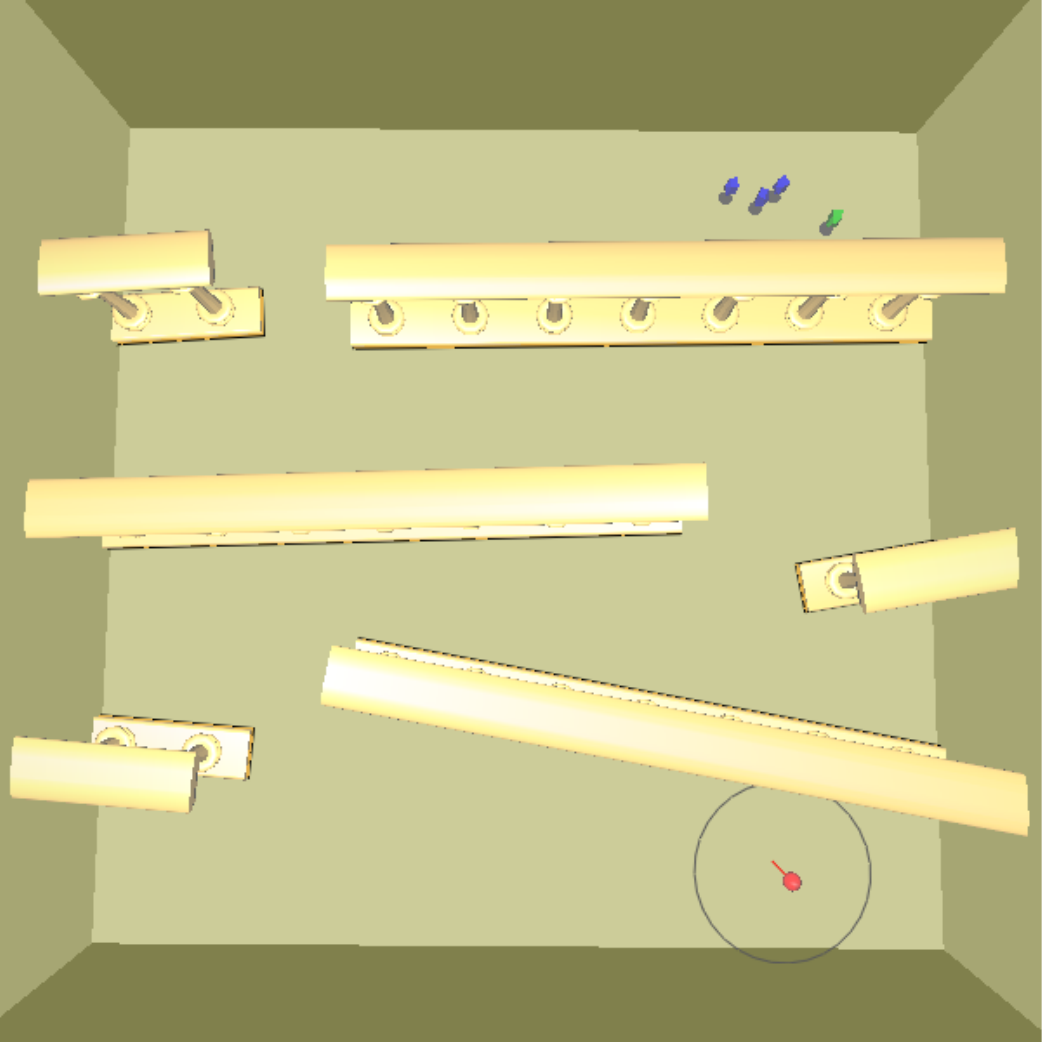}}
{\includegraphics[width = 0.23\textwidth]{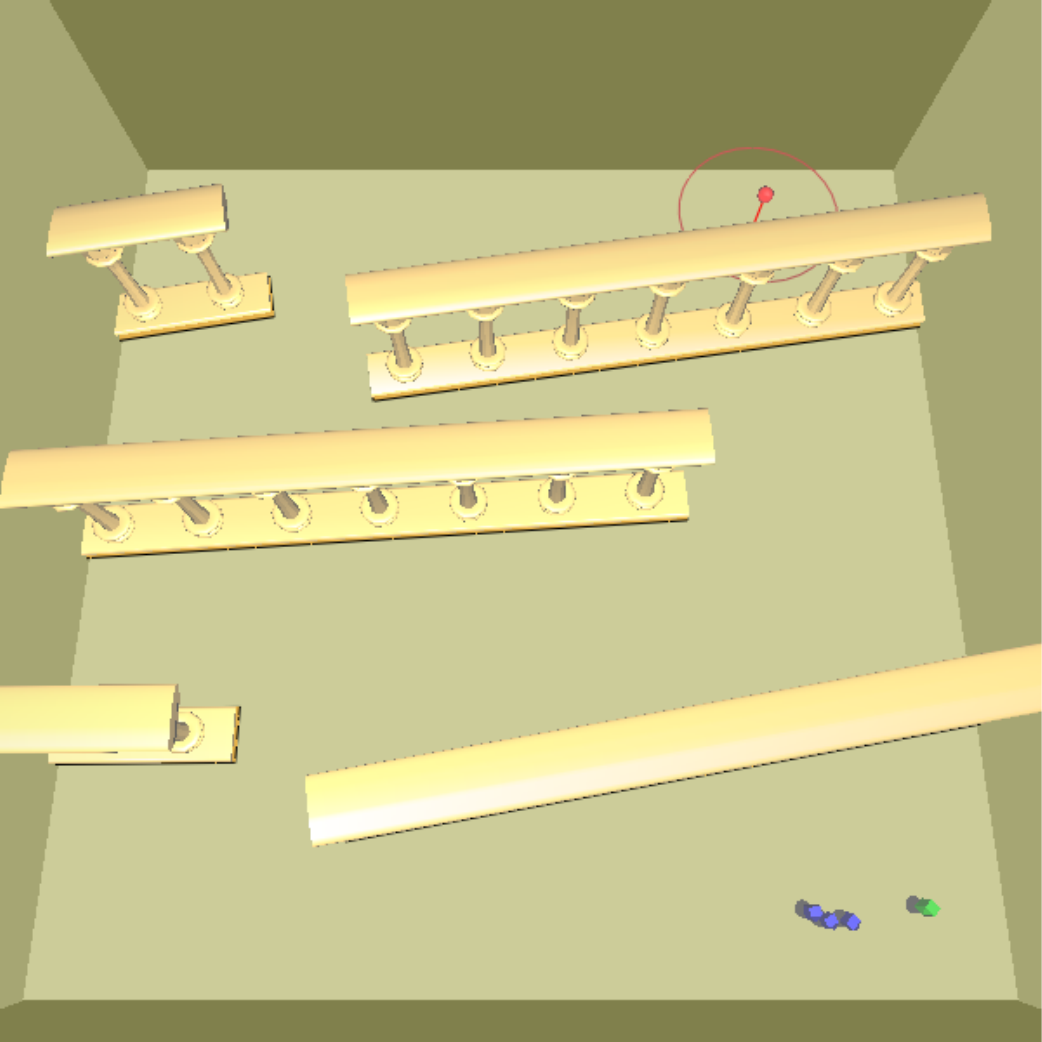}}
\caption{Two environments of three layers obstacles with U-turns and gaps. This environment is used for training and testing of more complex environments.}
\label{fig:layer}
\end{figure}

By combining the random and fixed obstacles, we consider randomly perturbed  obstacle patterns, i.e.,
the U-turn pattern and a small gap pattern illustrated in Figure~\ref{fig:layer}.

We create three-layer obstacle environments for training using these two patterns. 
There are at most three U-turns patterns and three gaps pattern using six fences in the left figure of Figure~\ref{fig:layer}. 
If we remove the shorter fences in the middle, it becomes a U-turn pattern in the figure on the right. 
For every fence, we set an initial position and angle. When we train the model, we add a clipped Gaussian noise to those default values. 
The range of position noise is between $-2.5$ and $+2.5$ pixel, and the range of the initial angle is between $-8^{\circ}$ and $8^{\circ}$. 
During the training, the shorter fences may appear or disappear randomly. 
For positions of sheep and goal, we set two positions and alternate between these two positions as the start and the goal points.


Recall that getting an accurate behavior model for sheep is nearly impossible, therefore, to make our shepherd more robust, we randomly change the group behavior parameters, namely the separation, alignment, cohesion, and fear coefficients, using  the clipped Gaussian distribution. To be more specific, the range of separation coefficient is between $0.5$ and $1.5$, the range of cohesion coefficient is between $3$ and $7$, the range of alignment coefficient is between $2$ and $4$, and the range of fear coefficient is between $13$ and $17$. The terminal time steps of each training episode is 8000. The average training time using perturbed obstacles is about 20 hours.


\subsection{Evaluating the Trained Shepherding Controller}
\label{sec:test}

\subsubsection{Representation}
The evaluate the trained model, we test the representation of the group (sheep) in the environments populated with random obstacles. To test, we use the same square environments, but randomly sample the goal in the environments. One-third of the goals are sampled around the corner, one-third are sampled around the edge and one-third are sampled around the center. We then randomly place the group in the environment and place the shepherd around the group in one of the eight directions. We design 300 test cases with different sizes of the group. 
Finally, we set different obstacle density from 0 to 0.3.

\begin{figure}[th]
\centering
{\includegraphics[width = 0.45\textwidth]{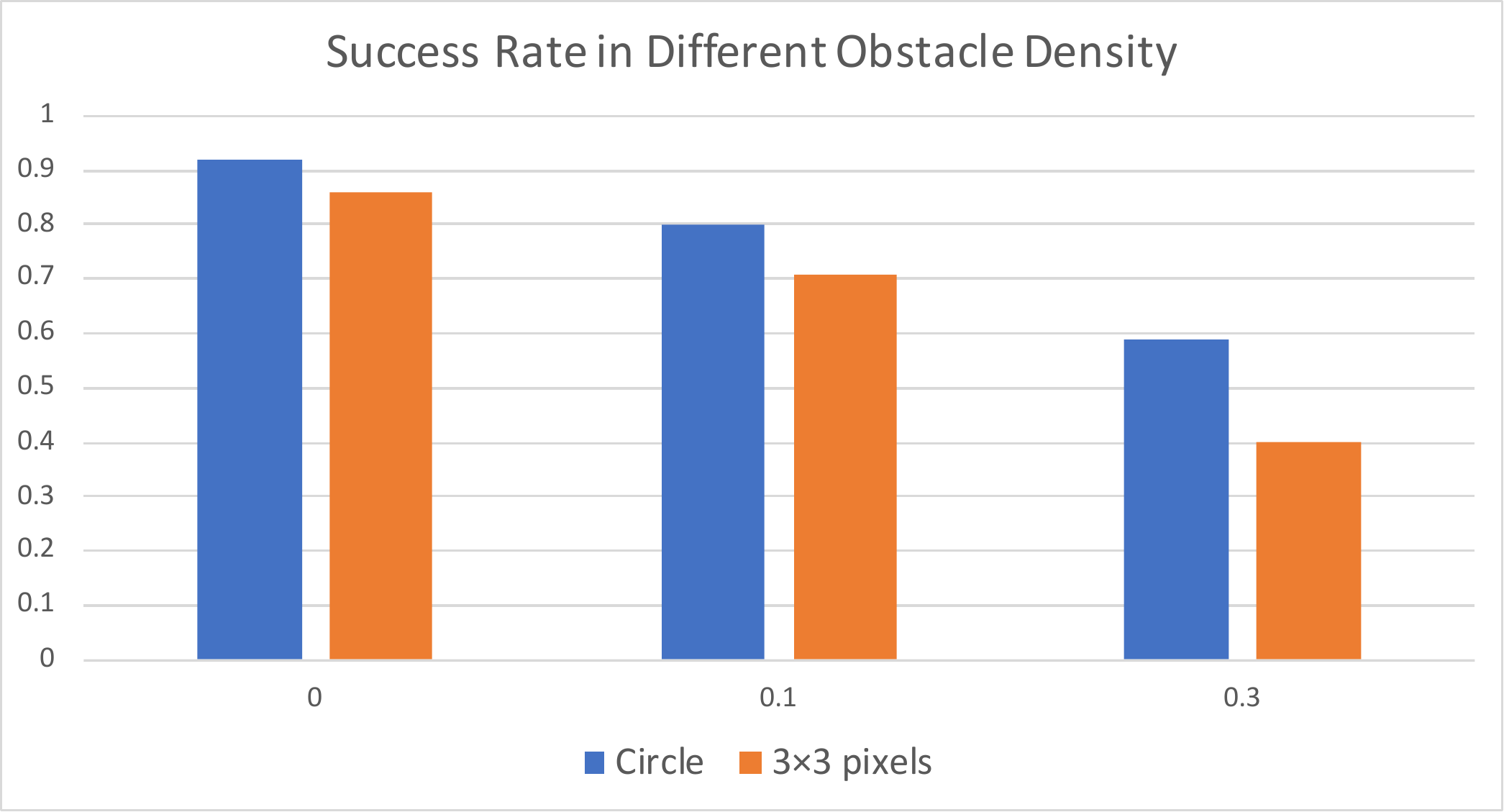}}
\caption{Success rates of the different data representation in different obstacle densities}
\label{fig:rep}
\end{figure}

We train two different models of three sheep in Section~\ref{sec:density} and obtain their success rate in different obstacle density. The terminal time steps of each testing episode is 4000. Figure~\ref{fig:rep} shows the result.


The results in figure~\ref{fig:rep} indicates that the circle shape of representation outperforms the pixels representations. We would use the circle shape in the rest of the experiments.

\subsubsection{Robustness}
In this experiment, we study the robustness of the trained model.
We use a fixed environment to study the robustness of the method subject to the noise in the environmental and behavioral parameters. 
More specifically, we change the behavior parameters of the sheep in two aspects: the fear force and group properties, i.e., the separation and cohesion forces.
In the first case, we study the effect of varying the \emph{fear force}. We change the fear force coefficient between $5$ and $25$ while the separation and cohesion coefficients are fixed at $1$ and $5$, respectively.
In the study of varying \emph{group property}, we change the separation coefficient from $0.5$ to $1.5$ and change the cohesion coefficient from $7.5$ to $2.5$ so that the agents are less likely to stay together on one end of the axis and more likely so on the other end. 
Therefore, the group will change from a concentrated group to a scatter group when the property changes from $0.5-7.5$ to $1.5-2.5$. 
In this second study, the fear force coefficient is fixed at $15$.

We compare the model trained by the proposed method  with two rule-based methods, namely the simple-rule method and complex-rule method \cite{lien2004shepherding}. 
When applying the simple rule, the shepherd moves straight behind the group to a new steering point. The complex rule method uses a side-to-side moving policy around the steering point behind the group and also positions itself proactively to prepare the sheep for the upcoming turning. Both methods are capable of merging the separated sheep back to a single group.

\begin{figure}[th]
\centering
{\includegraphics[width = 0.23\textwidth]{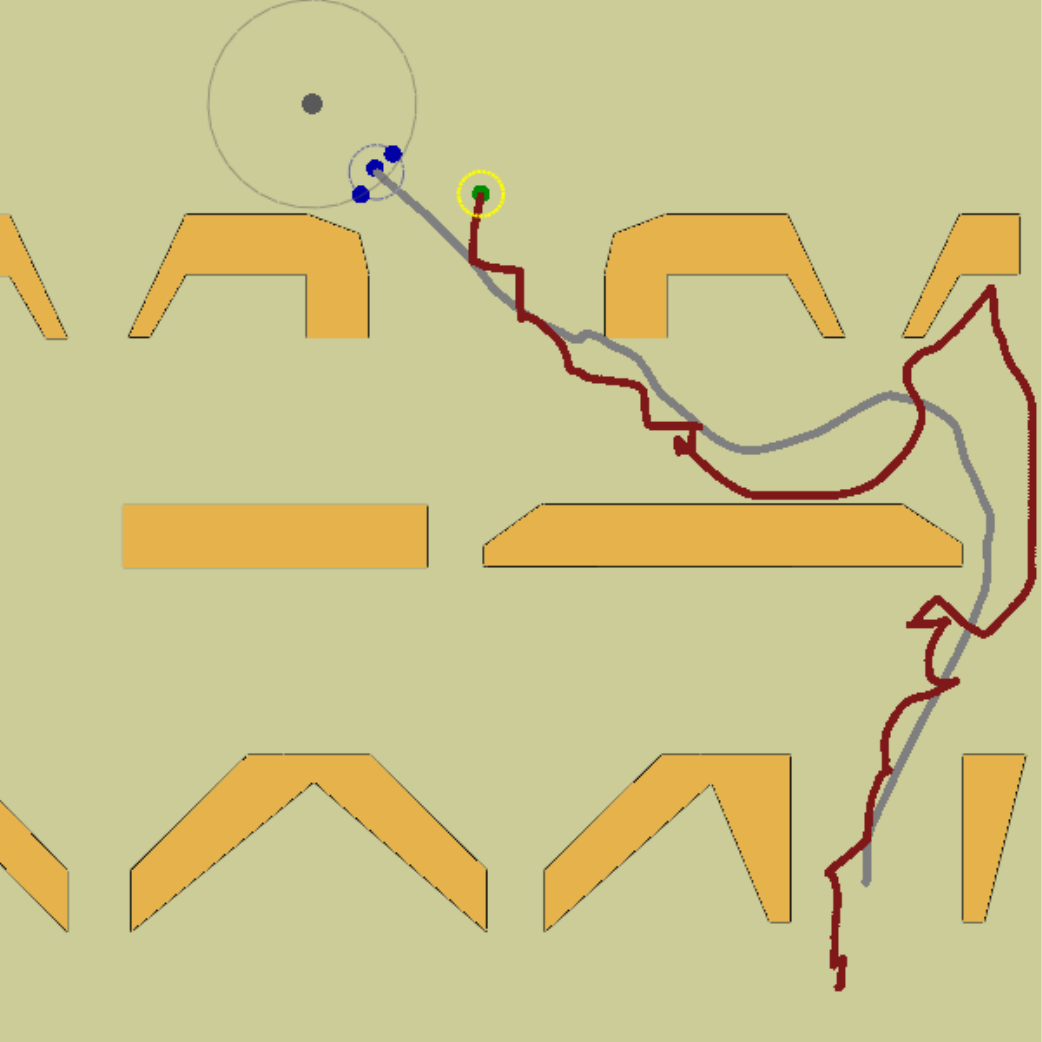}}
{\includegraphics[width = 0.23\textwidth]{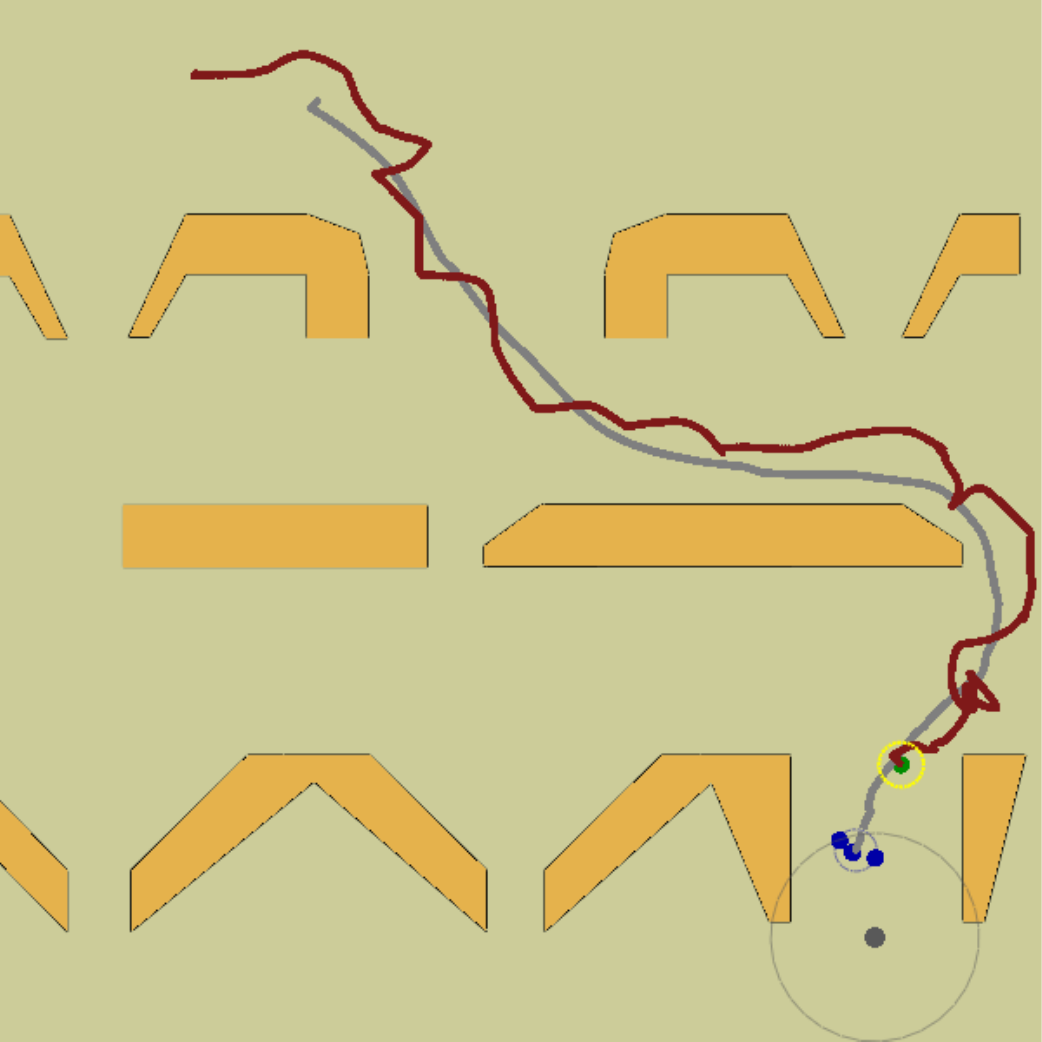}}
\caption{The filter environment, shepherd path (red) \& sheep (center) path (grey) }
\label{fig:filter}
\end{figure}

Figure~\ref{fig:filter} shows a trajectory of the shepherd that successful herd a group of three sheep. Since back and forth paths of filter environment are quite different, we split the filter environment into filter-0 and filter-1 environments. Left part in Figure~\ref{fig:filter} is filter-0 environment, the goal is on the top. Right part in Figure~\ref{fig:filter} is filter-1 environment. It is easy to see herding in filter-1 environment is easier than heading in filter-0 environment.

We have three aspects to study the robustness: success rate, completion time and path length. In these experiments, we conduct and collect data from 500 runs for each method in each environment and the terminal time steps of each testing episode is 5000.


\begin{figure}[th]
\centering
{\includegraphics[width = 0.48\textwidth]{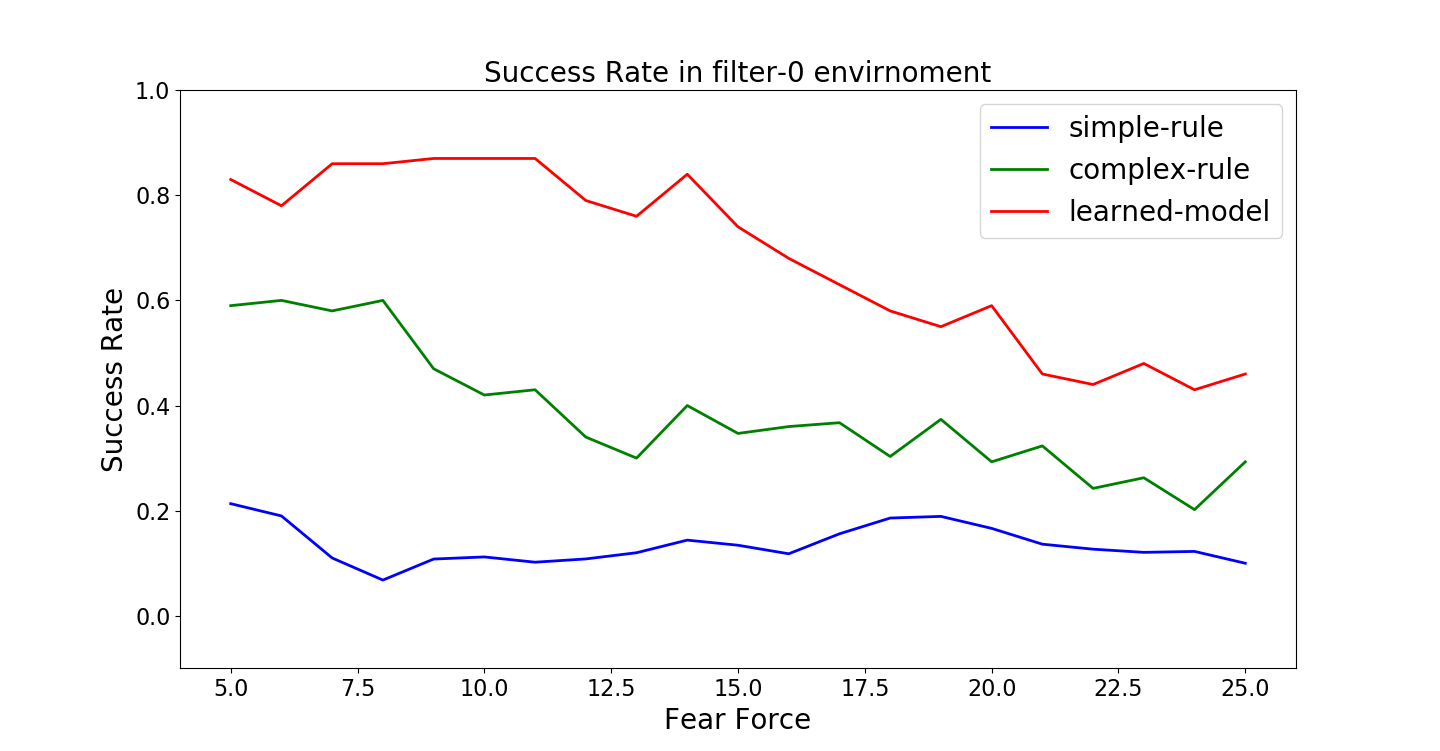}}
{\includegraphics[width = 0.48\textwidth]{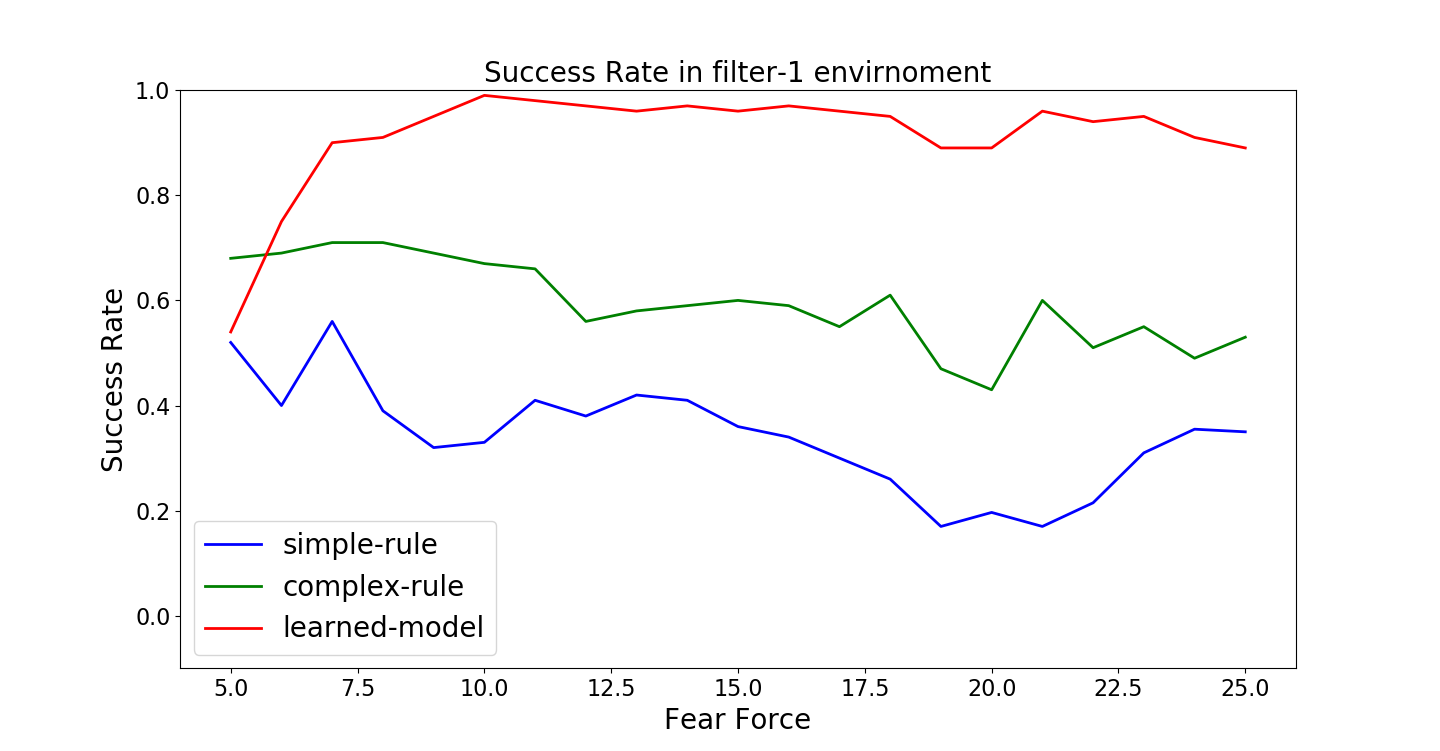}}

\caption{Success rate in different levels of fear force in the filter environment. High fear force models easily scared agents (e.g., geese) that are more difficult to control. 
The learned model is trained with fear coefficient $15$.} 
\label{fig:filter-fear}
\end{figure}

\begin{figure}[th]
\centering
{\includegraphics[width = 0.48\textwidth]{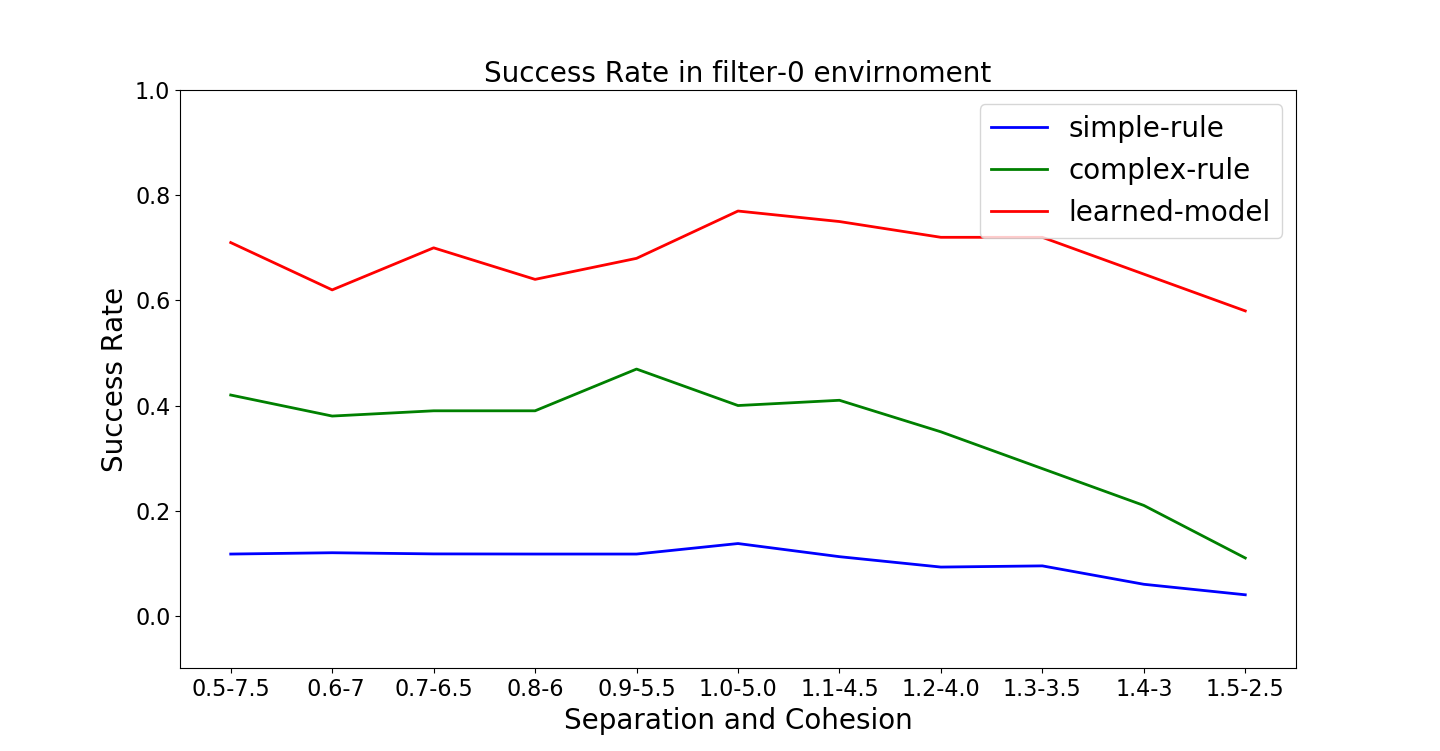}}
{\includegraphics[width = 0.48\textwidth]{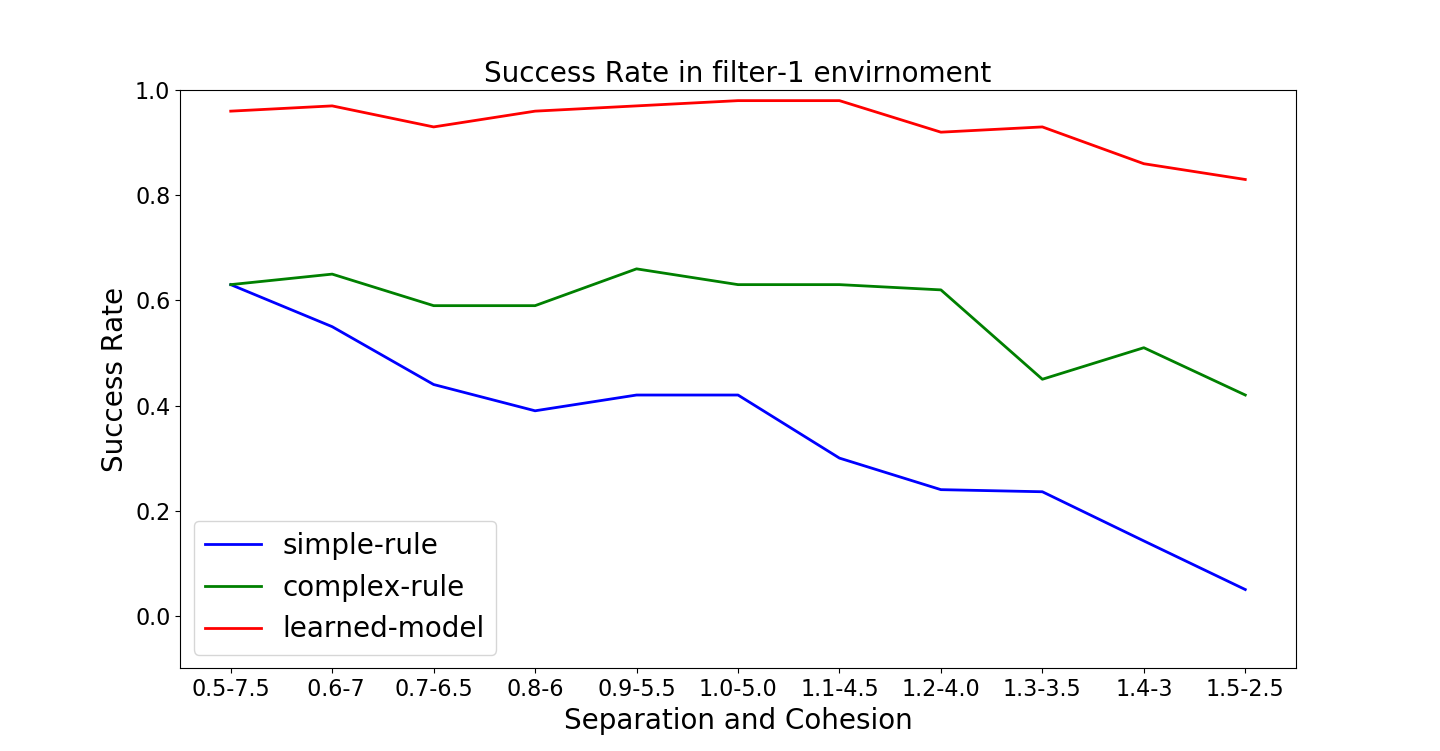}}
\caption{Success rates in different levels of separation and cohesion in the filter environment. The $x$-axis covers a range of group  behaviors from low separation and high cohesion (e.g. sheep) to high separation and low cohesion (e.g., cattle). }
\label{fig:filter-sc}
\end{figure}

\textbf{Success Rate}. 
In Figures~\ref{fig:filter-fear} and~\ref{fig:filter-sc}, the proposed model maintains a higher success rate compared to the other methods even when the 
controlled group becomes more difficult to herd.

In Figure~\ref{fig:filter-fear}, even when we train the controller using the fear force at value $15$, the model shows a high success rate for a large ranges of fear coefficient from $5$ to $25$.
In Figure~\ref{fig:filter-sc}, since the range of separation force coefficient is between $0.5$ and $1.5$, and the range of cohesion force coefficient is between $3$ and $7$ during the training, the success rate of learned model maintains the same high-level success rate compared to the other two rule-based methods. 



\textbf{Completion Time}. In addition to the success rate, we are interested in knowing the quality of the path created by the learned model.
To measure the quality using the  time needed to complete a herding task, we study two types of measurements: (1) the time of successful cases and 
(2) the time of both successful and failed cases.

\begin{figure}[h]

\centering
{\includegraphics[width =0.48\textwidth]{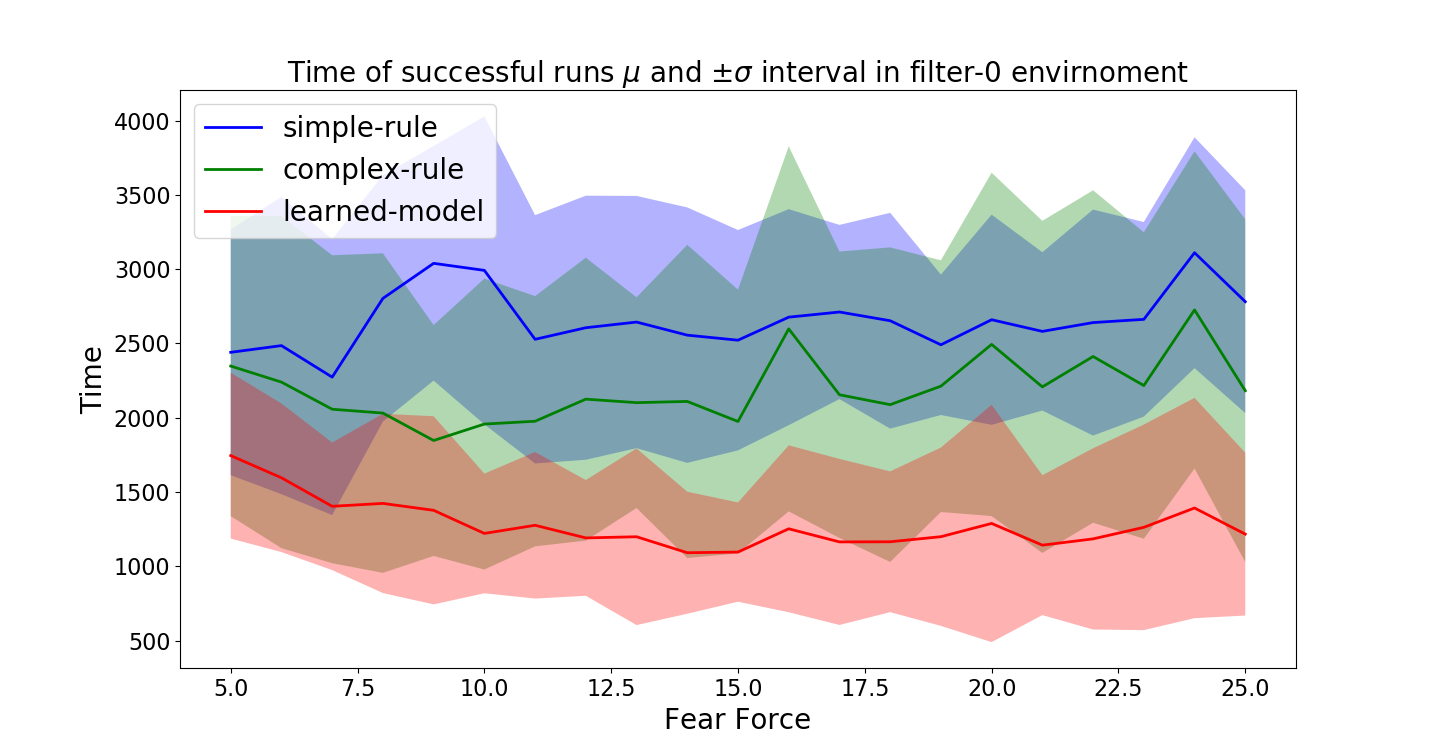}}
{\includegraphics[width = 0.48\textwidth]{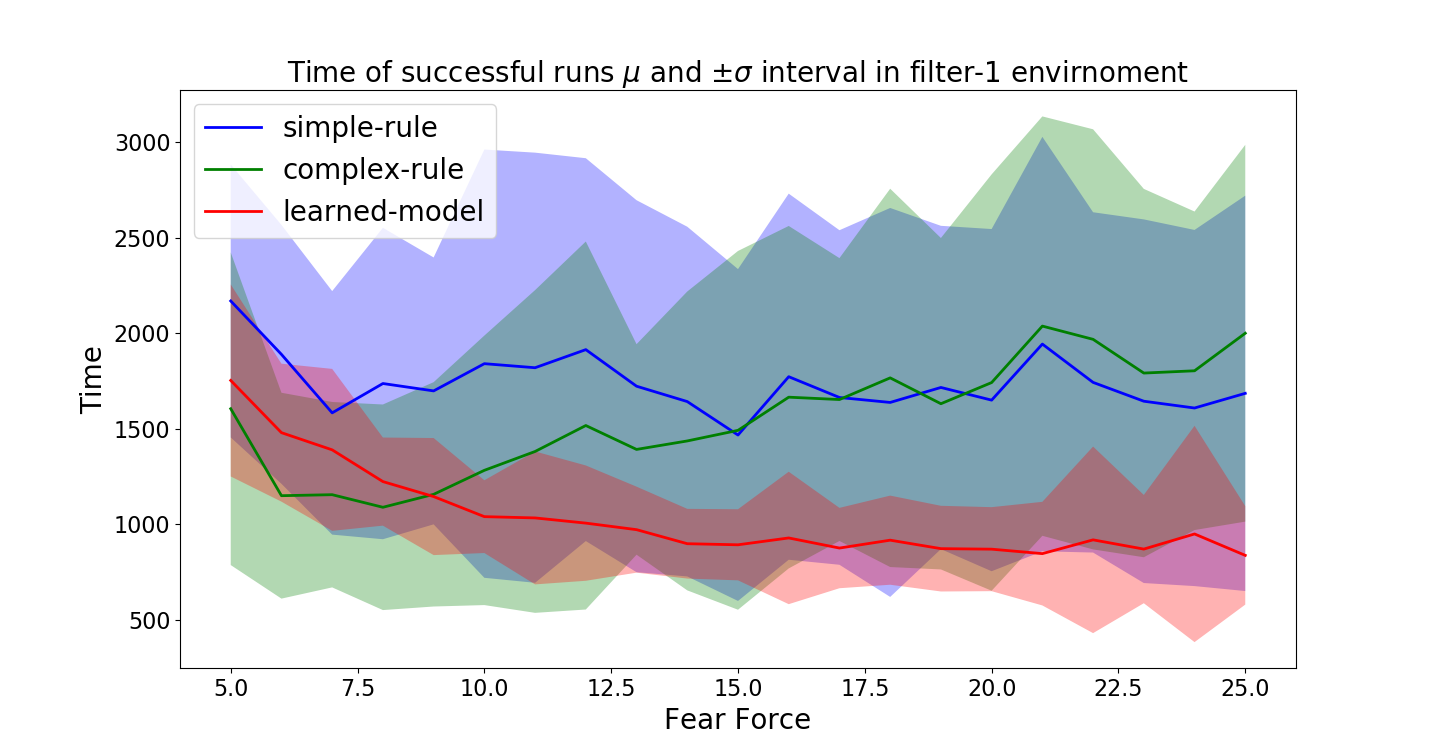}}
\caption{Completion time of the successful runs for different levels of fear coefficient in the filter environment.  }
\label{fig:filter-f-stime}

\end{figure}

\begin{figure}[h]
\centering
{\includegraphics[width = 0.48\textwidth]{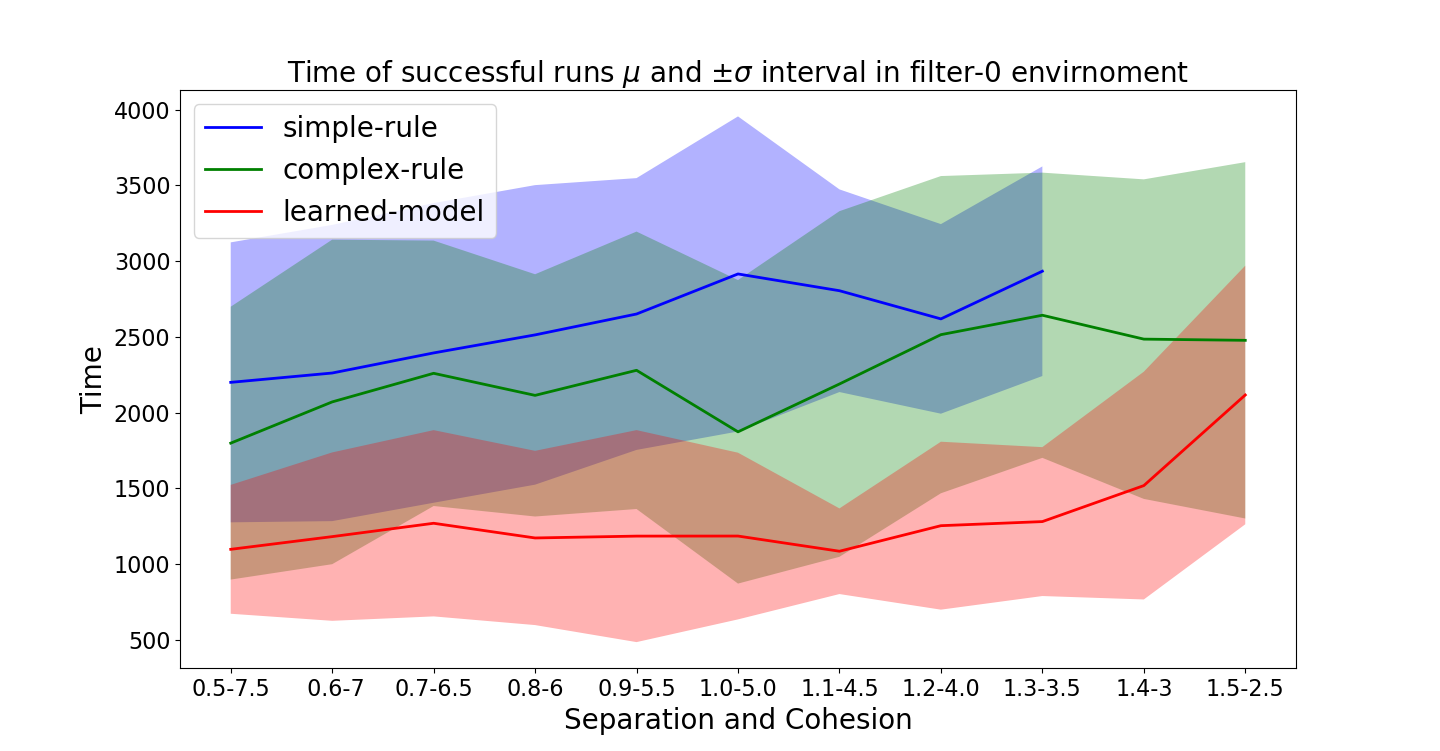}}
{\includegraphics[width = 0.48\textwidth]{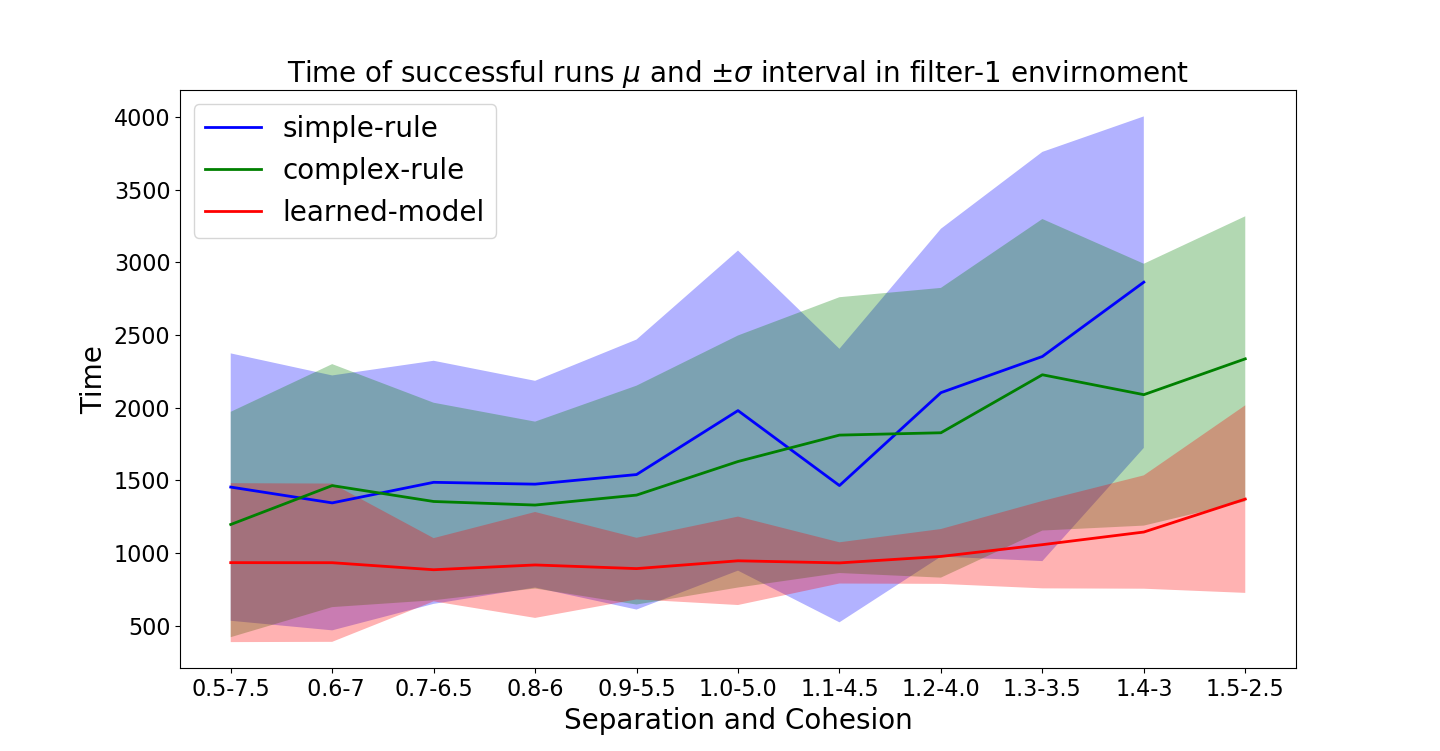}}
\caption{Completion time of the successful runs for different levels of separation and cohesion in the filter environment.  }
\label{fig:filter-sc-stime}
\end{figure}

Figures~\ref{fig:filter-f-stime} and~\ref{fig:filter-sc-stime} show the completion time of the successful cases for different levels of parameters in the filter environments. 
In these figures, the learned model spends less time herding than the other two rule-based methods. Simultaneously, the standard deviation of completion time is also the smallest among all methods. The figures show that the learned model completed the herding task as soon as possible, and it is robust to maintain this quality.  

Note that in Figure~\ref{fig:filter-sc-stime}, the time of simple-rule method is empty in some parts, which is due to the number of data is less than 30 (No statistically significant) or the success rate is 0.

\begin{figure}[h]
\centering
{\includegraphics[width =0.48\textwidth]{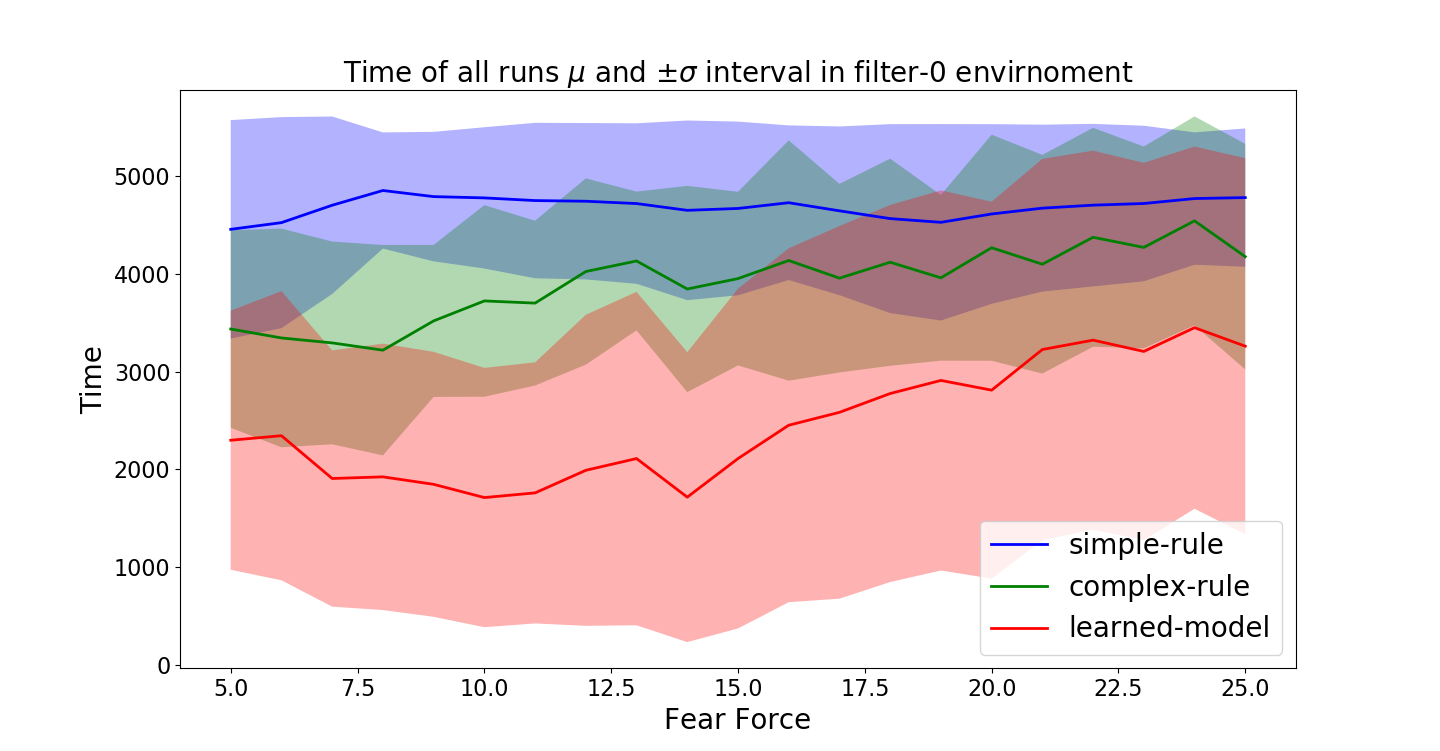}}
{\includegraphics[width = 0.48\textwidth]{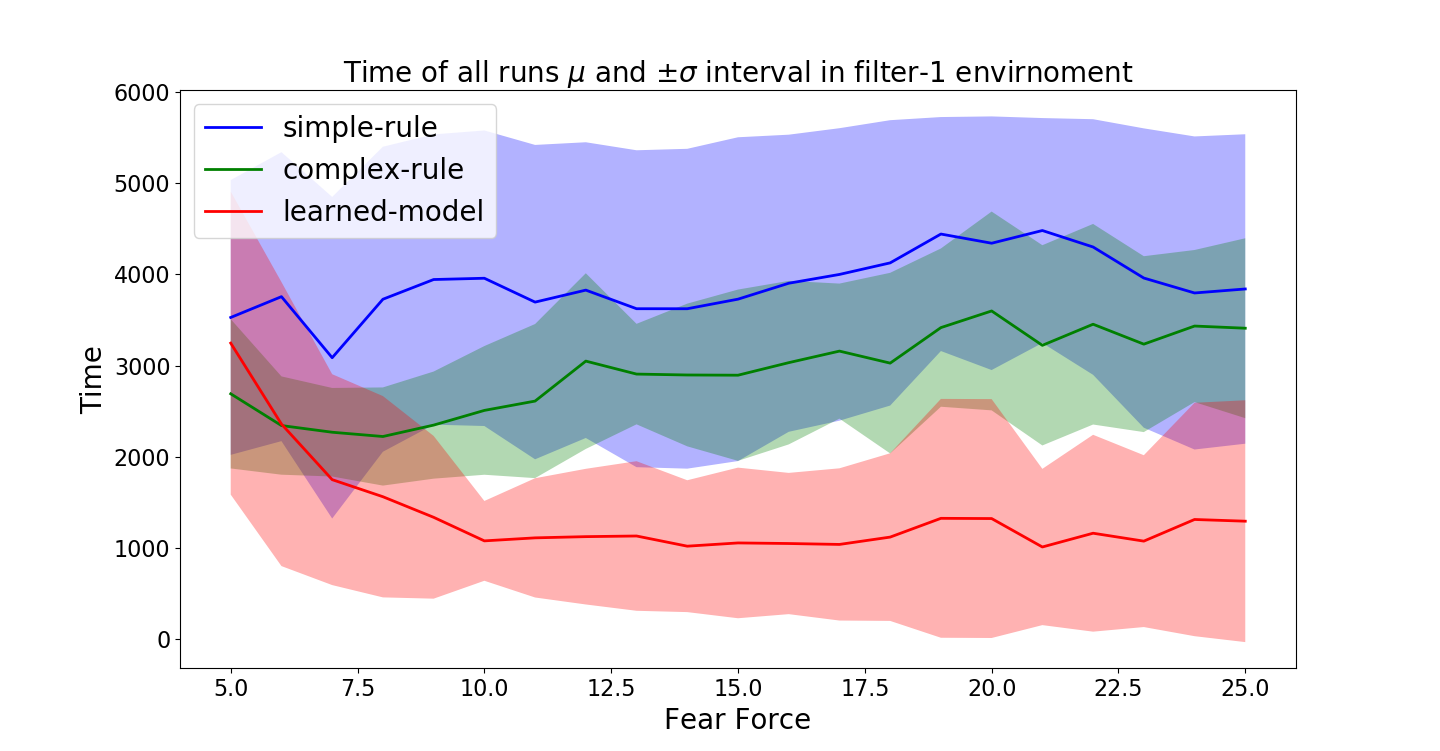}}
\caption{Time of all runs for different levels of fear coefficient in filter environment.  }
\label{fig:filter-f-atime}
\end{figure}

\begin{figure}[h]
\centering
{\includegraphics[width = 0.48\textwidth]{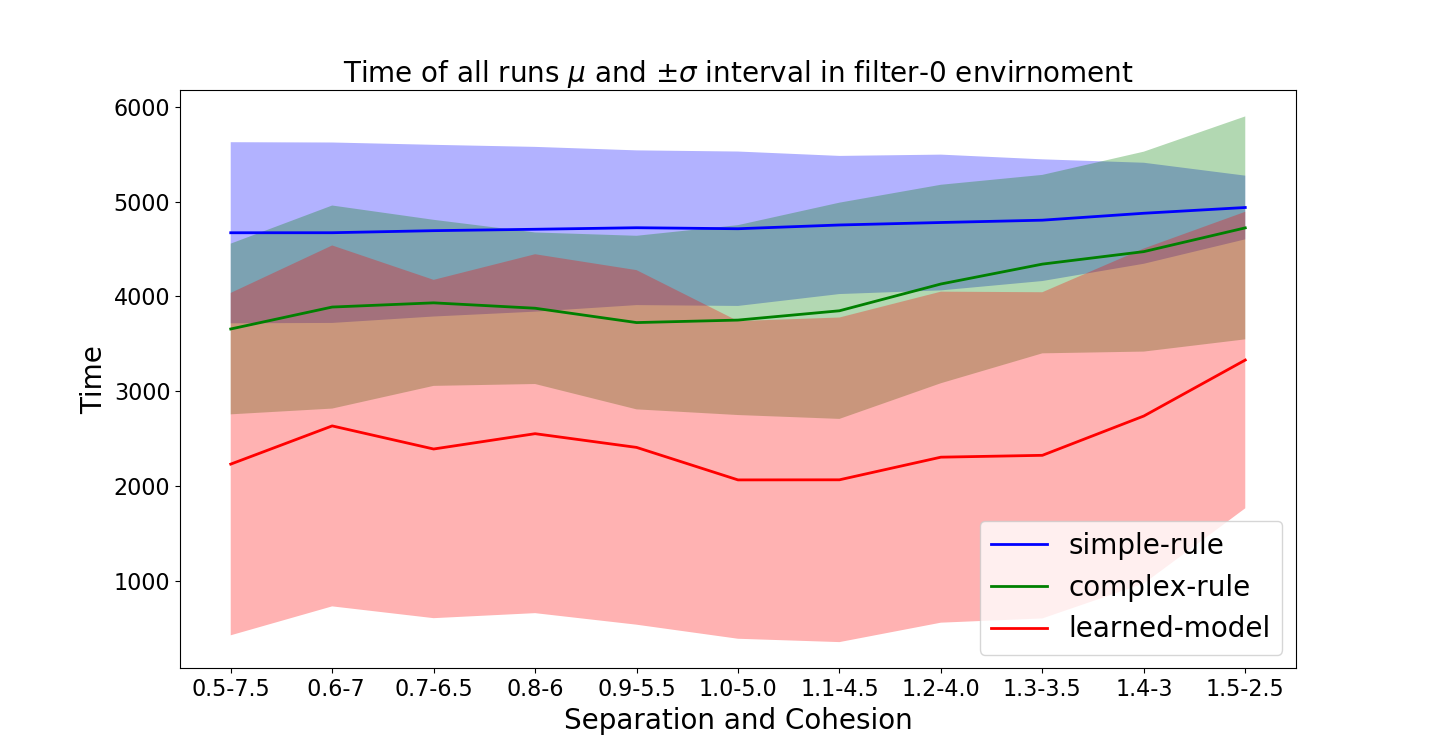}}
{\includegraphics[width = 0.48\textwidth]{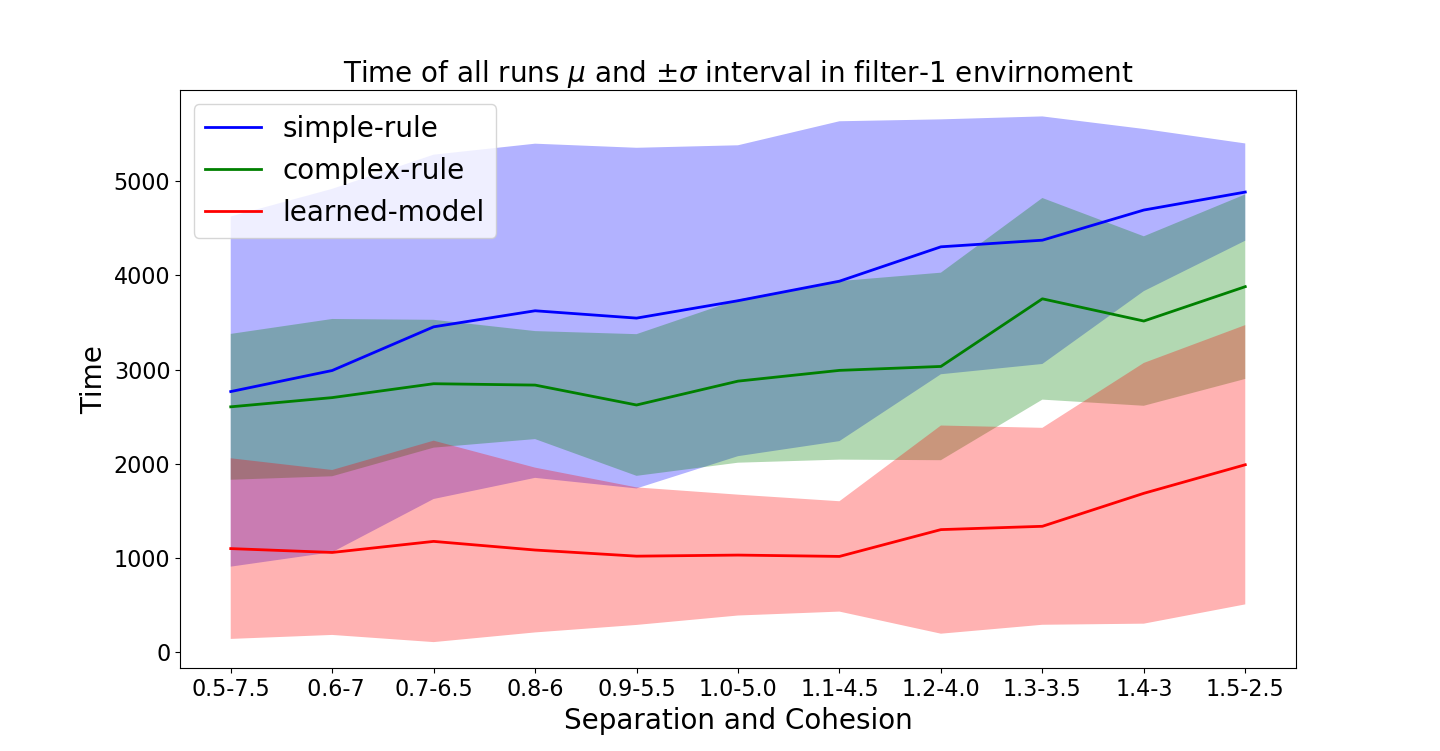}}
\caption{Time of all runs  for different levels of separation and cohesion in filter environment.  }
\label{fig:filter-sc-atime}
\end{figure}

For the completion times of all (failed and successful) cases, we  set the time to the maximum allowed time (5000 in this experiment) for the failed cases. 
Figures~\ref{fig:filter-f-atime} and~\ref{fig:filter-sc-atime} show the completion times for different levels of group behavior parameters in filter environment. 
As given in these figures, the learned model still spends the least time among all methods.
Since the maximum allowed time is much larger than the average time of success runs, the standard deviation of the learned model method is high. 

\begin{figure}[th]
\centering
{\includegraphics[width =0.48\textwidth]{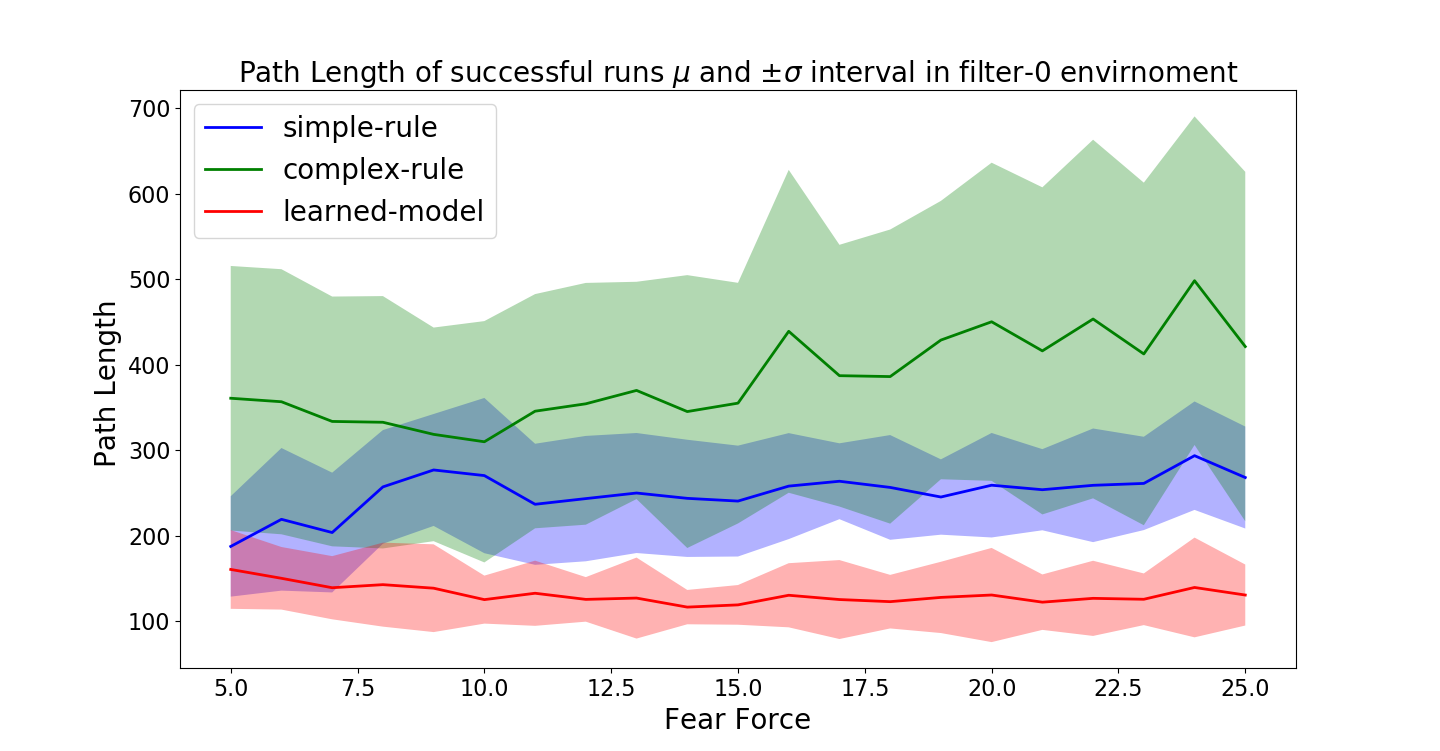}}
{\includegraphics[width =0.48\textwidth]{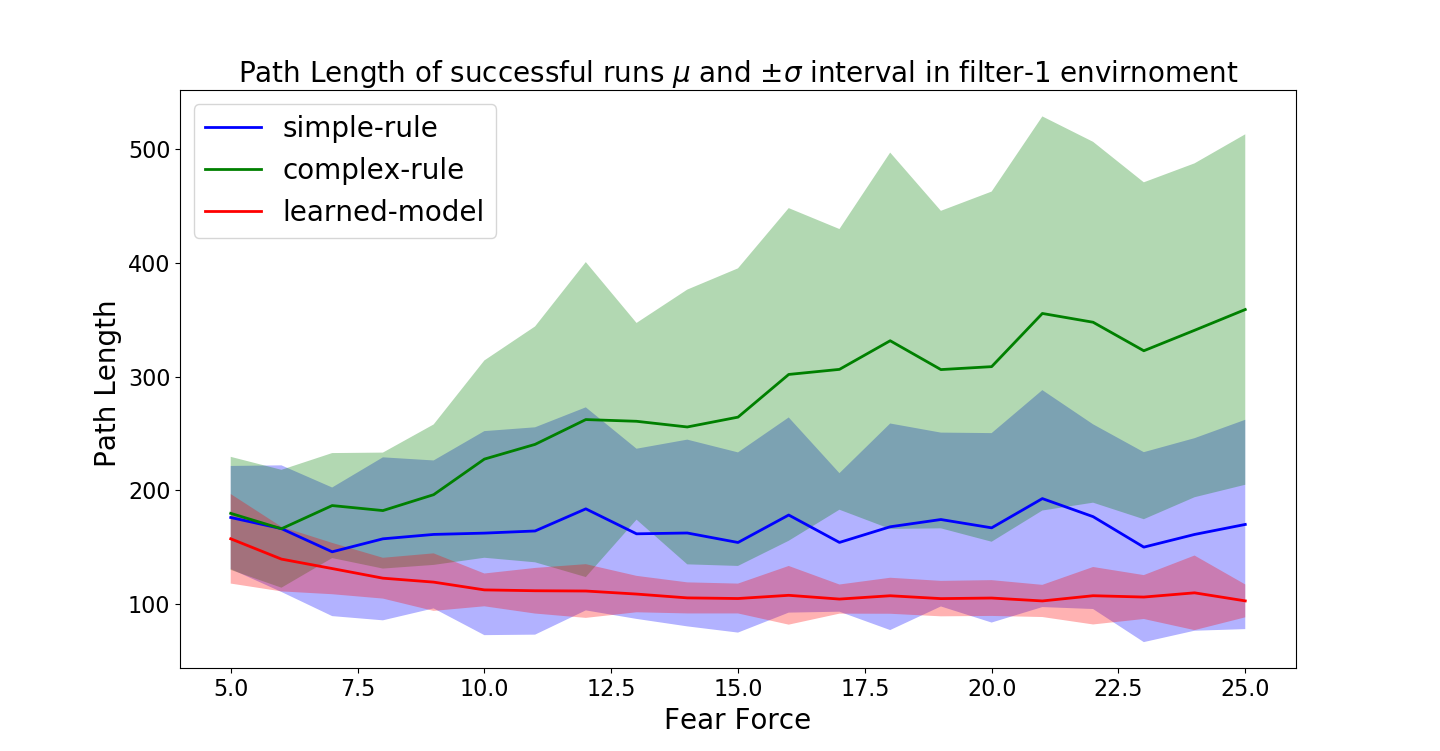}}
\caption{Path length of the successful runs for different levels of fear coefficients in the filter environment. }
\label{fig:filter-f-spath}
\end{figure}

\begin{figure}[h!]
\centering
{\includegraphics[width =0.48\textwidth]{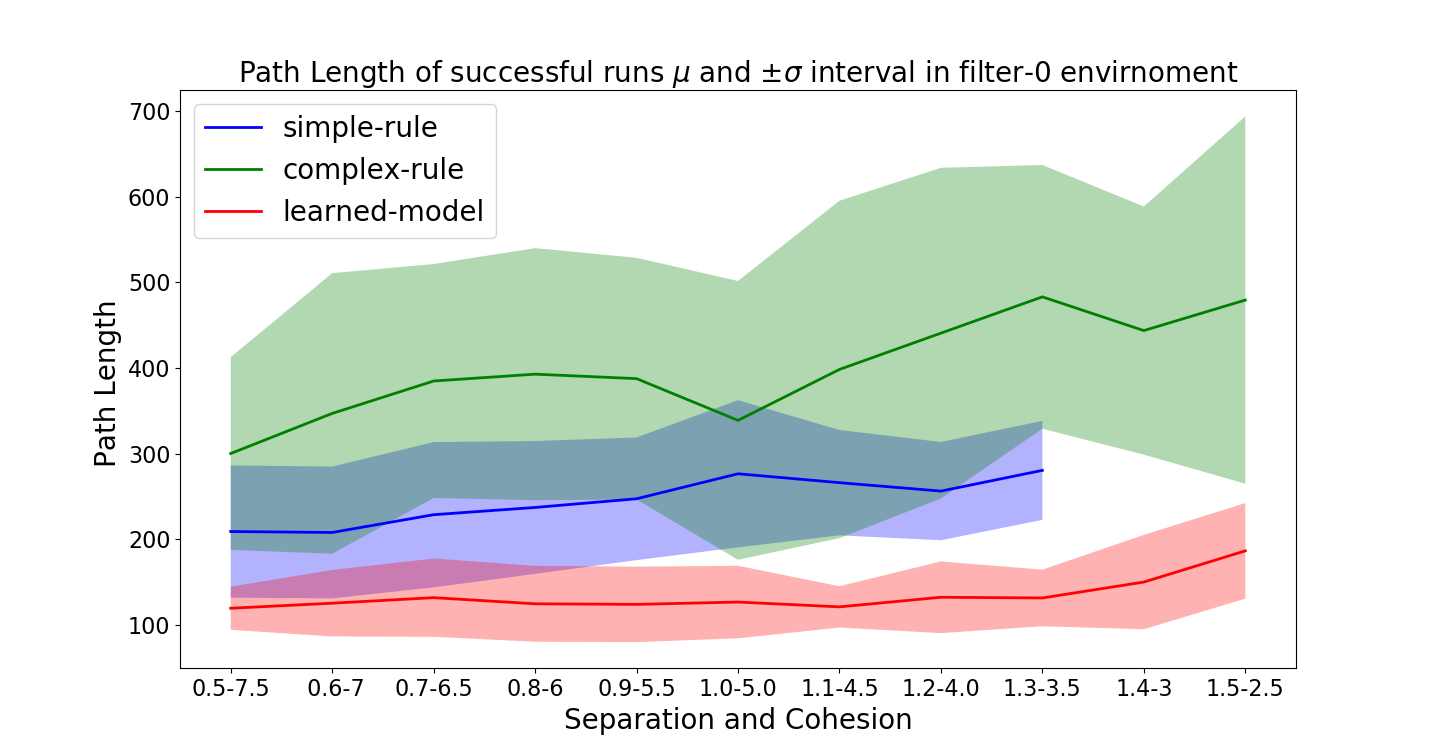}}
{\includegraphics[width =0.48\textwidth]{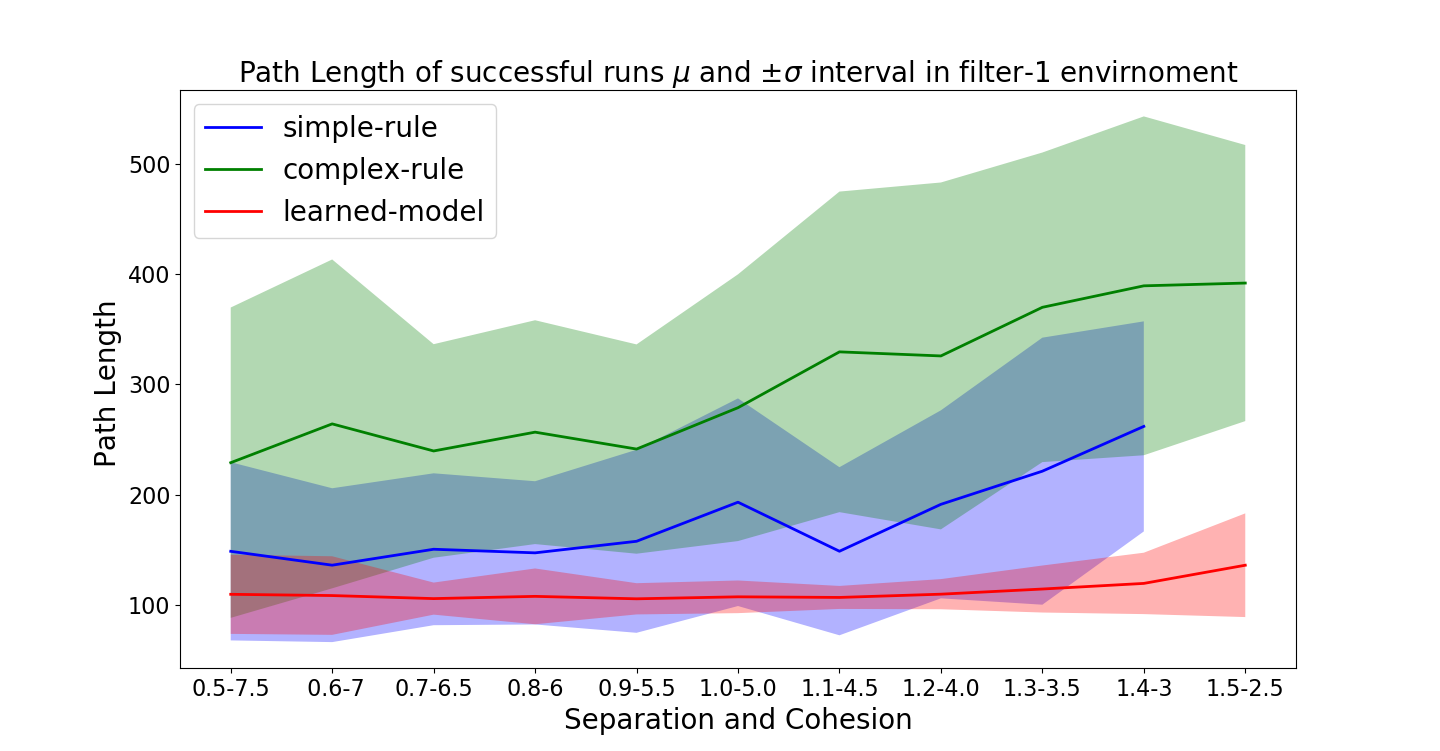}}
\caption{Path length of the successful runs for different levels of separation and cohesion in  the filter environment. }
\label{fig:filter-sc-spath}
\end{figure}

\textbf{Path Length}.
We also use the path length to measure the quality of path. 
For the path length, we only study path length of success cases because the shepherd often does not move when it failed to herd.
Figures~\ref{fig:filter-f-spath} and~\ref{fig:filter-sc-spath} show path length of successful cases for different levels of group behavior parameters in the filter environments. From these figures, we find that the shepherd controlled by the learned model moves significantly less compared to that controlled by the rule-based methods. This property is useful when considering the energy cost.
We also note that the standard deviation of mean path length with the learned model is smaller than other rule-based algorithms. This indicates that the learned model is robust to maintain its efficiency even when the group becomes more difficult to control.

\begin{figure*}[h]
\centering
{\includegraphics[width=0.23\textwidth]{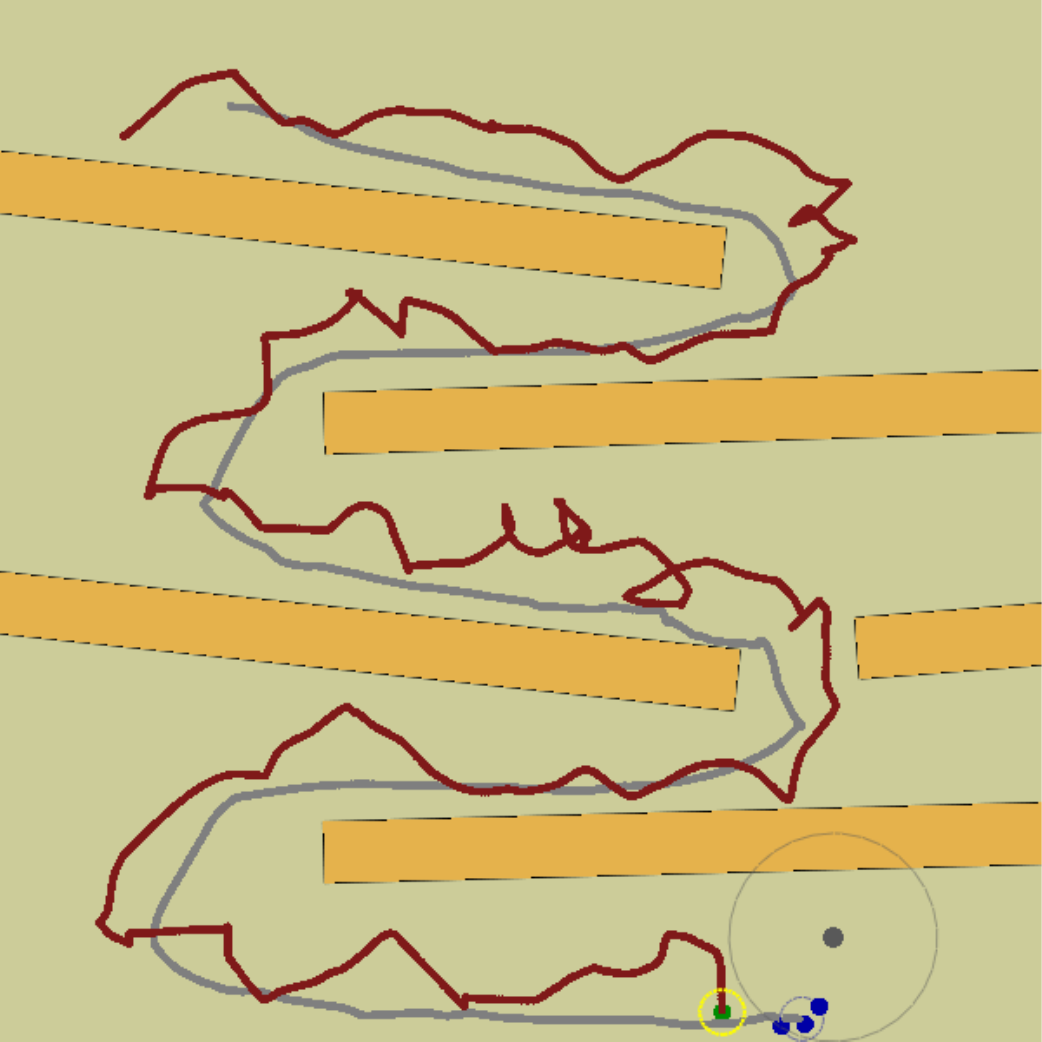}}
{\includegraphics[width=0.23\textwidth]{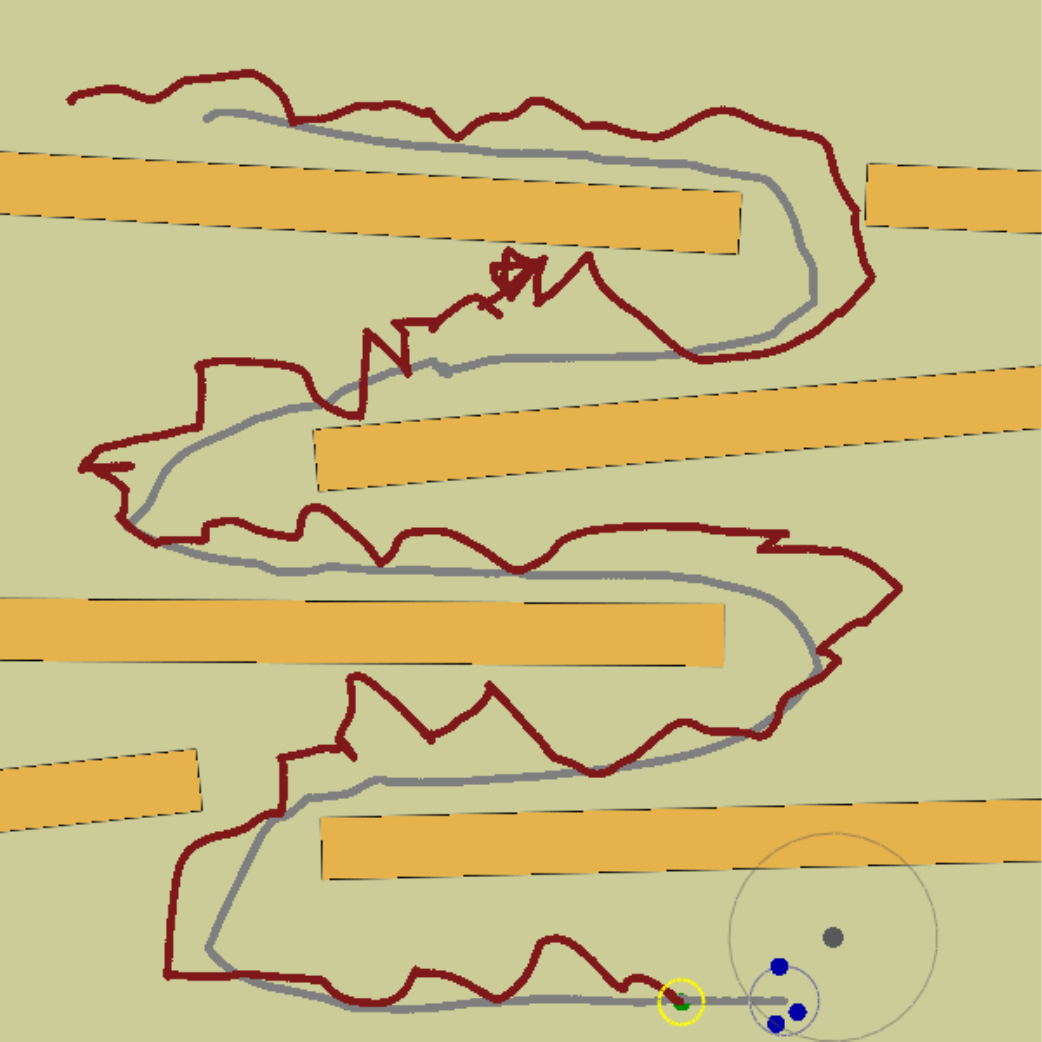}}
{\includegraphics[width=0.23\textwidth]{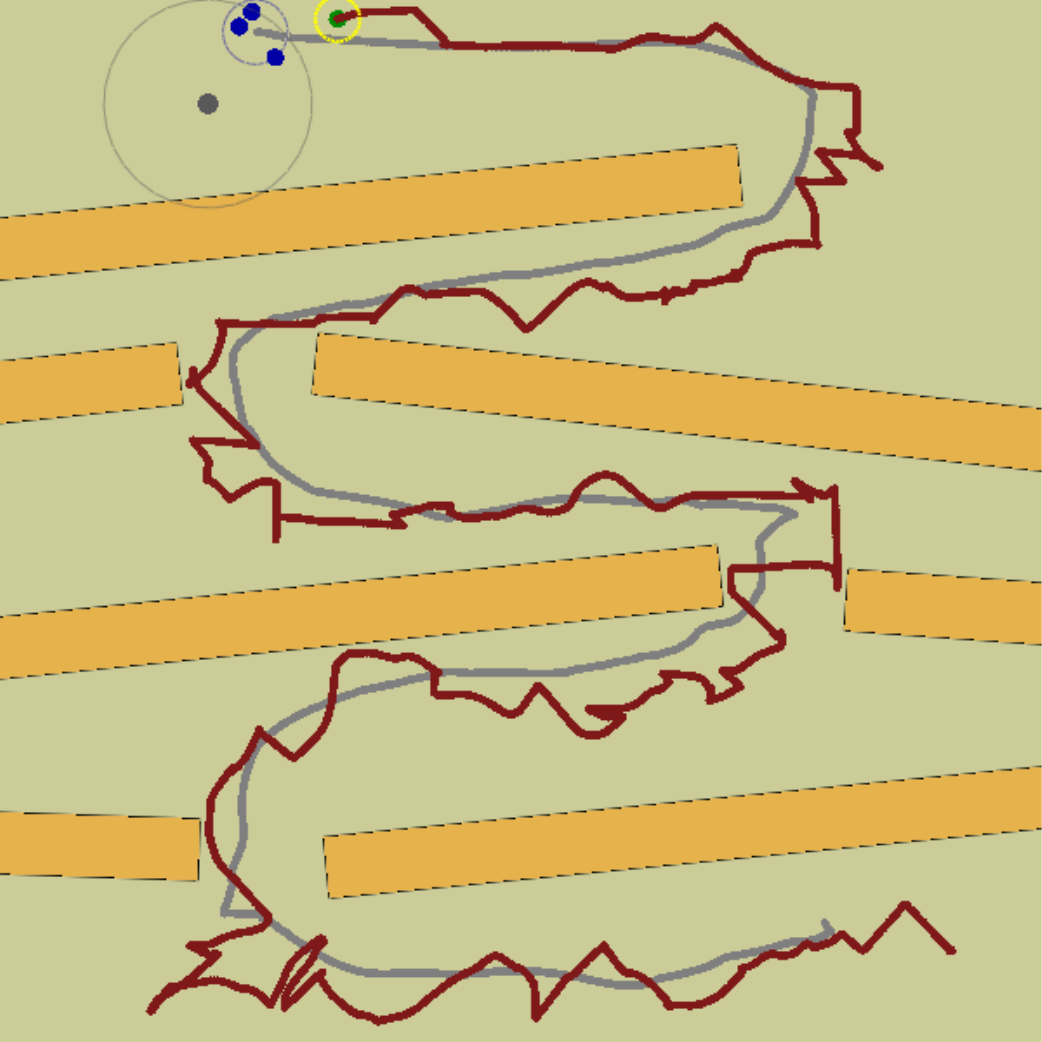}}
{\includegraphics[width=0.23\textwidth]{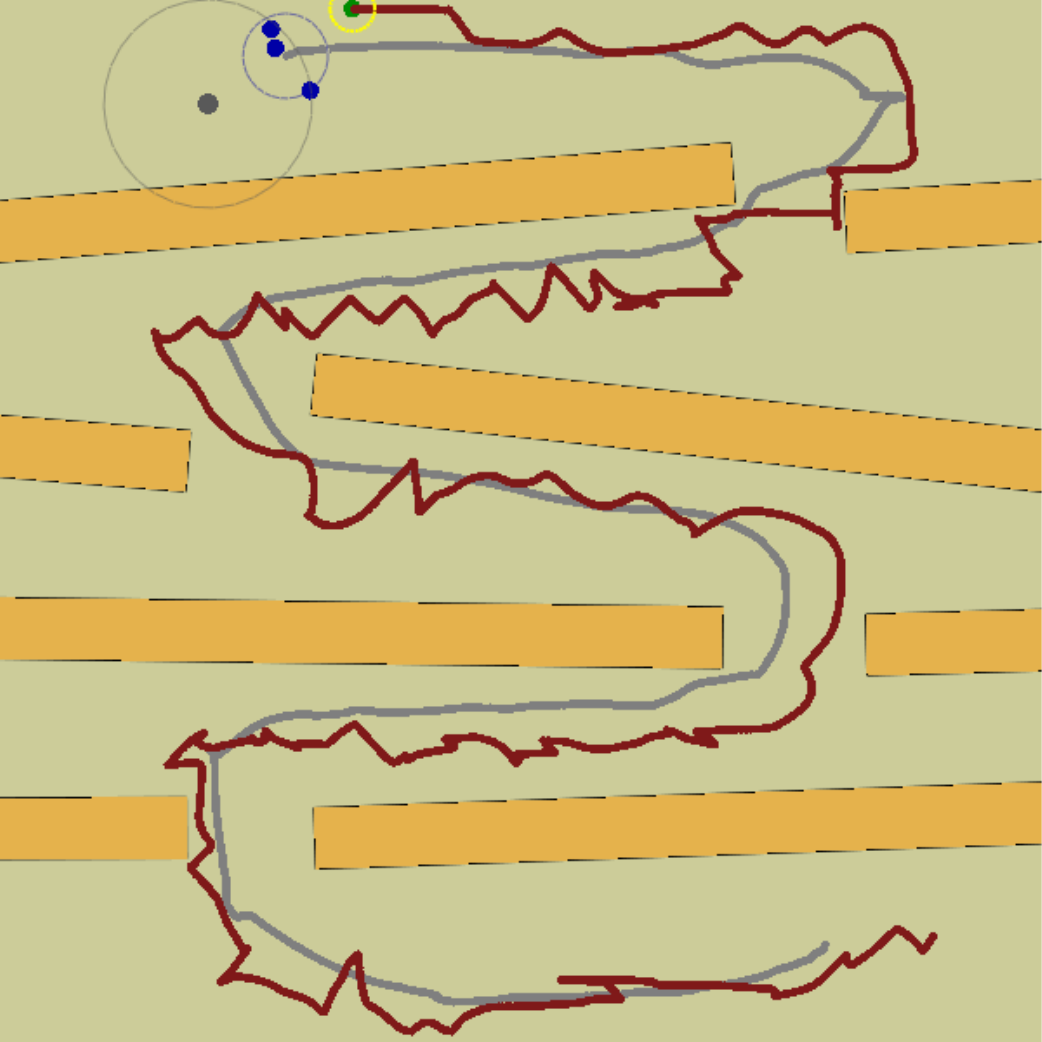}}
\caption{The four-layer obstacles with U-turns and gaps. From left to right: one gap, two gaps, three gaps and four gaps. In all environments, the shepherd's path are in red, and the sheep  paths are in grey.}
\label{fig:test}
\end{figure*}

\subsubsection{Success rate with perturbed obstacles}
In these experiments, we study the success rate of the learned model in perturbed obstacles as described in Section~\ref{sec:training_perturb}. This will allow us to study the performance of the learned model in more realistic scenarios in which sensors are noisy and the environmental model may contains uncertainty. 
To test the learned model, we randomly choose parameters for the group behavior in a range identical to those used in the training and randomly choose different patterns to develop three layers and four layers obstacles. 

To evaluate the success rate of the learned model, we design three variations of the three-layer obstacle environment: one gap, two gaps, and three gaps. See Figure~\ref{fig:layer}. These environments have 
one gap in one of three U-turns randomly, or 
two of three U-turns randomly or gaps in all the U-turns. 
A set of similar but more difficult environments containing four layers 
with one to four gaps is shown in  Figure~\ref{fig:test} which also illustrates the successful paths of shepherd and the sheep center.


\begin{figure}[th]

\centering
{\includegraphics[width = 0.45\textwidth]{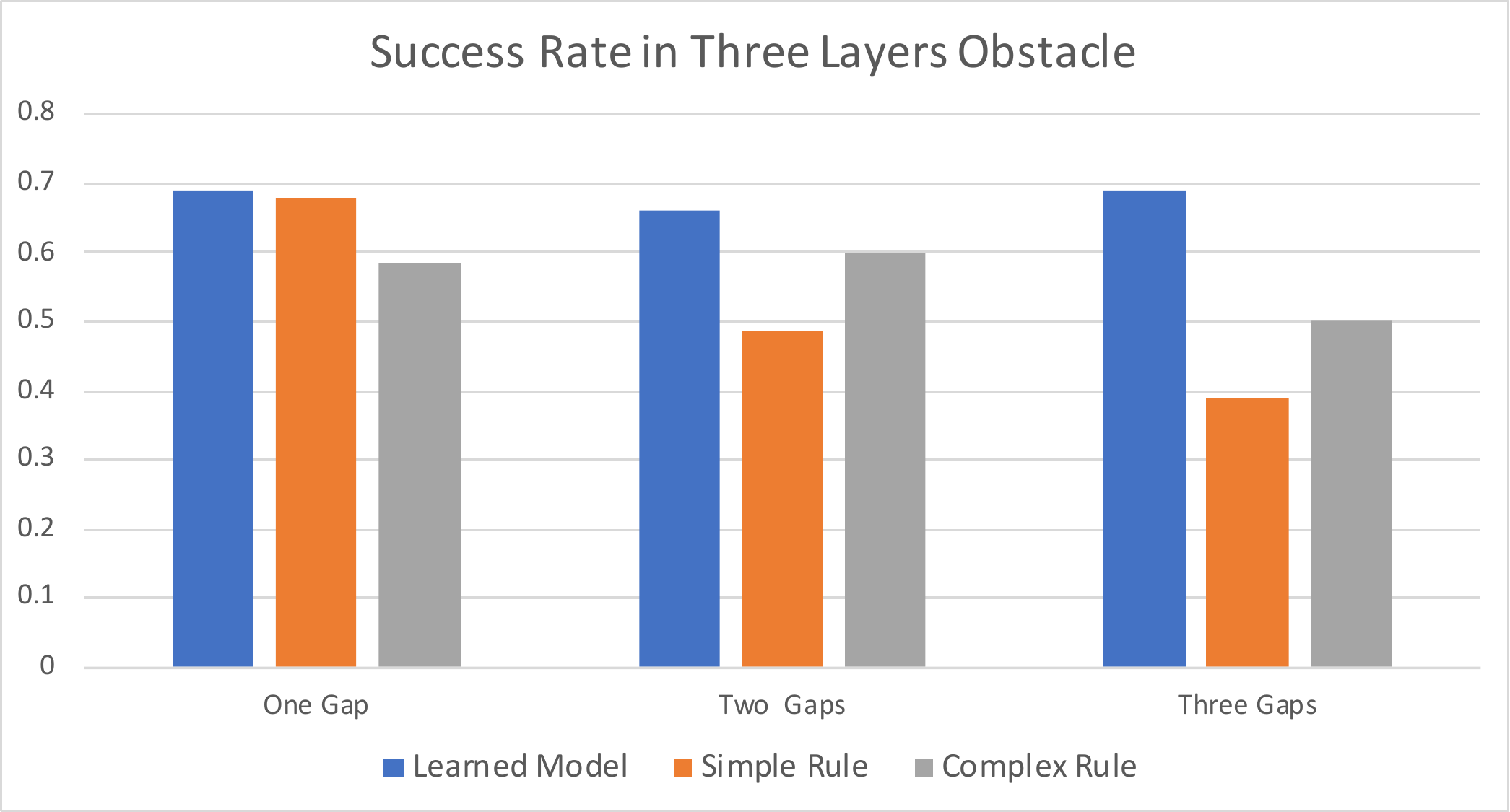}}
{\includegraphics[width = 0.45\textwidth]{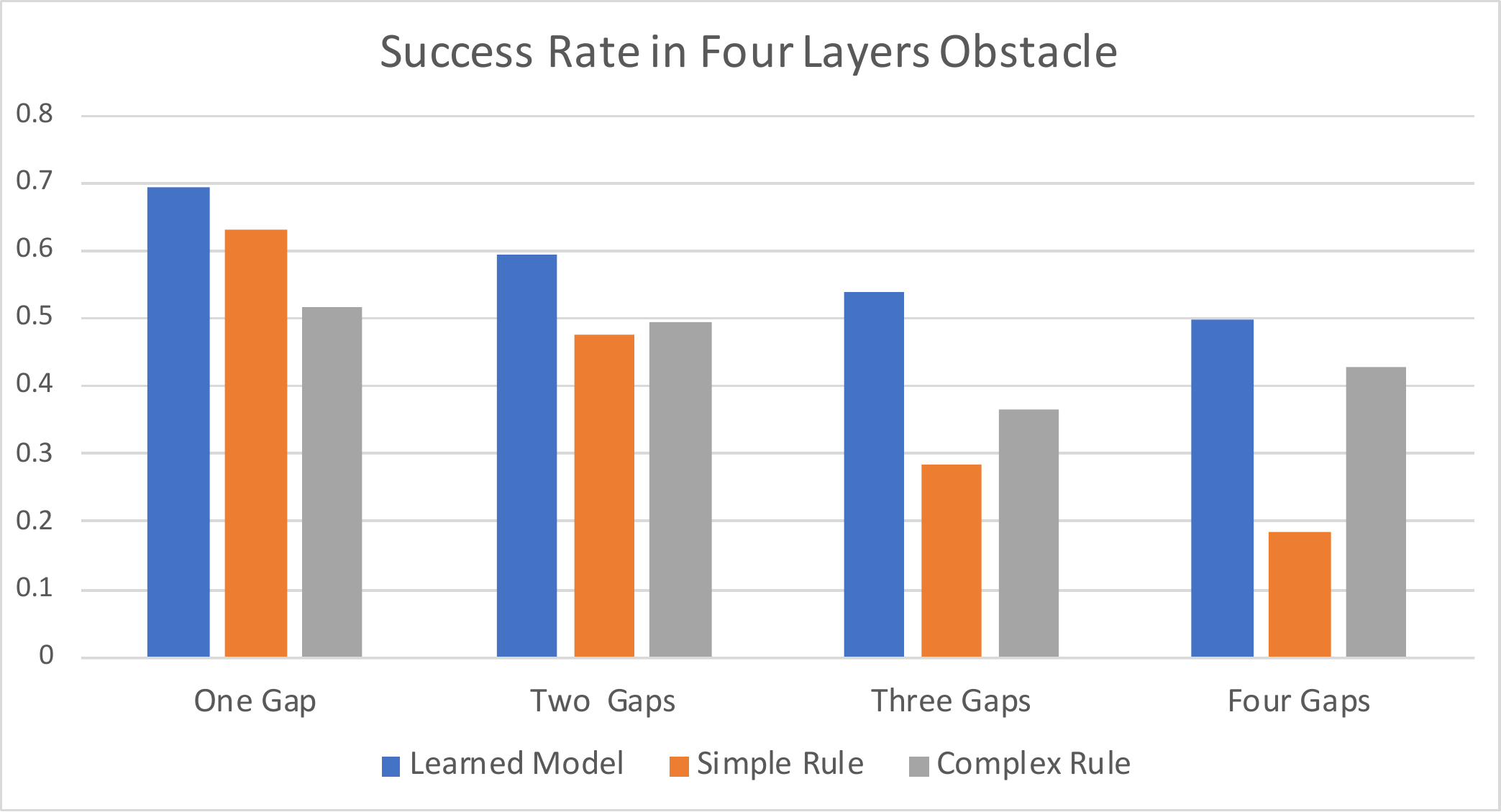}}
\caption{Success rates in randomly generated (top) three-layer obstacles and (bottom) four-layer obstacles. }
\label{fig:3s}
\label{fig:4s}
\end{figure}

With these environments, 
we conduct 100 experiments for each method in each environment, and the terminal time steps of each testing episode is 10000. Therefore, the total of 2100 experiments are conducted. 
As given in the Figure~\ref{fig:3s}, the proposed model has at least 20\% higher in success rate compared to the rule-based methods when there are more gaps in the layers. 

Form Figure~\ref{fig:3s} top part, the success rate of the learned model is around 70\% for all three types because we trained the model in the three-layer environment and it has at least 40\% higher than the rule-based methods for three gaps environment.

From Figure~\ref{fig:4s} bottom part, the test environments are more difficult than the training environments, so the success rate decreased, but the success rate of the learned model remains at least 20\% higher than the rule-based methods. 

\subsubsection{Robustness  of herding model in fixed environment trained using perturbed obstacles}

In this section, we use the model trained with perturbed obstacles to study the success rate and path quality, which is measured using the completion time and path length.
We conduct experiments in the most difficult four-layer obstacles environment with four U-turns and four gaps as our test environment.
The obstacles are not perturbed in all trails of this experiment. The terminal time steps of each testing episode is 10000.

\begin{figure}[th]
\centering
{\includegraphics[width = 0.48\textwidth]{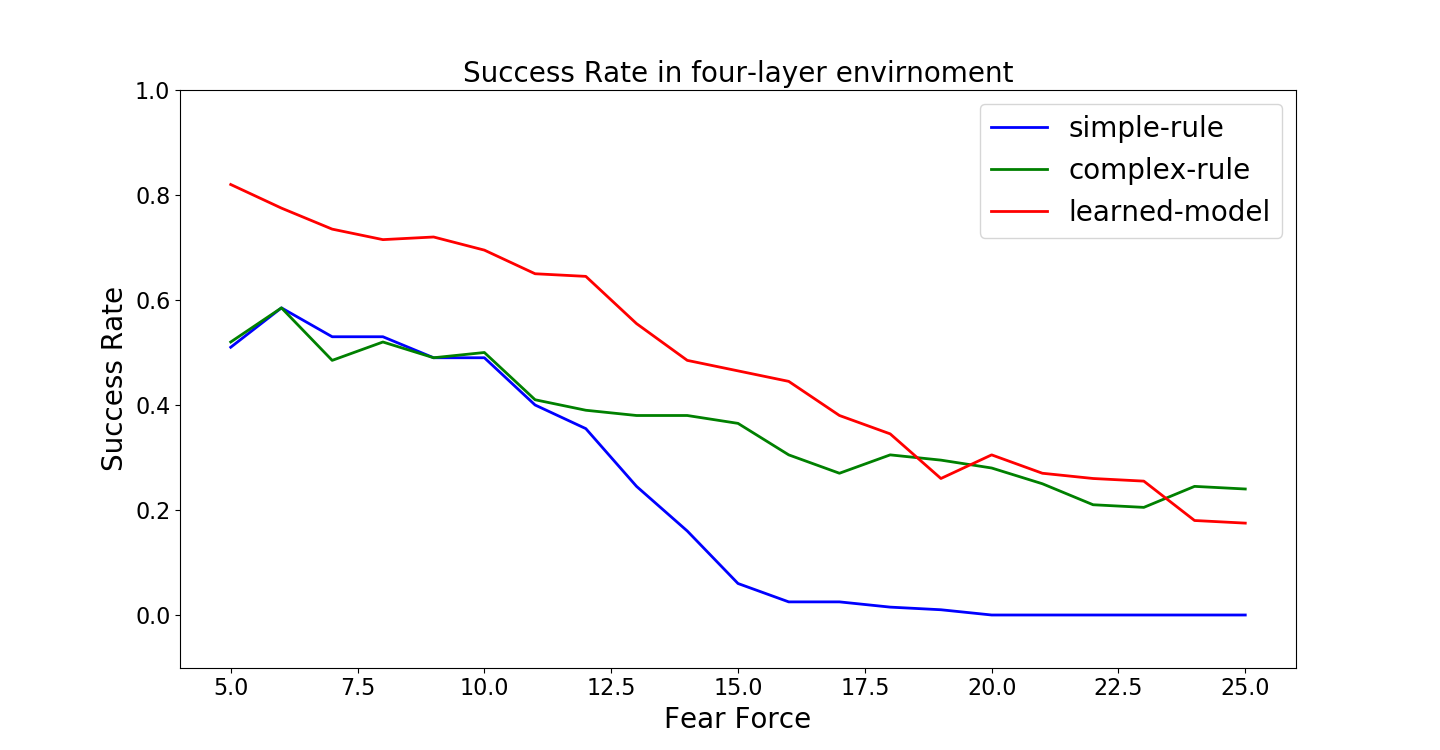}}
\caption{Success rates in different levels of fear force in fixed four-layer obstacle environment. High fear force models easily scared agents (e.g., geese) that are more difficult to control. }
\label{fig:4ts-fear}
\end{figure}

\begin{figure}[th]
\centering
{\includegraphics[width = 0.48\textwidth]{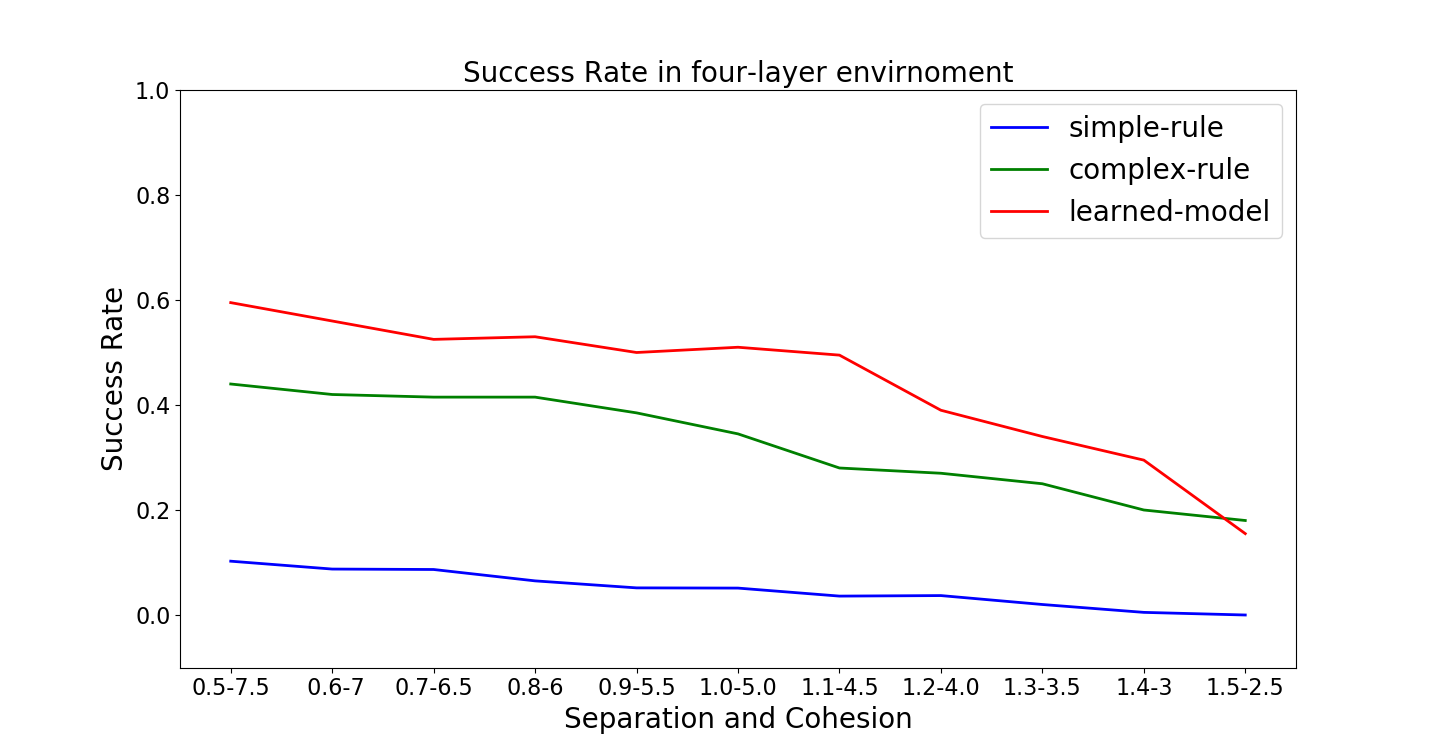}}
\caption{Success rates in different levels of separation and cohesion in fixed four-layer obstacle environment. The $x$-axis covers a range of group behaviors from low separation and high cohesion (e.g. sheep) to high separation and low cohesion (e.g., cattle). }
\label{fig:4ts-sc}
\end{figure}

\textbf{Success Rate}.
Figures~\ref{fig:4ts-fear} and~\ref{fig:4ts-sc} show the success rates. In Figure~\ref{fig:4ts-fear}, when the fear force coefficient increases, the success rate of all methods decreased, but the success rate of the learned model decreased gently compared to the simple rule method, and remains higher than the complex rule method as well as simple rule method. 
Note that Figure~\ref{fig:4ts-fear} partially explains that the success rate of the learned model in Figure~\ref{fig:4s}  is around 50\% in the environments with four gaps. 
Because the range of fear force coefficient in testing phase is between $13$ and $17$, which is a large value. 
If we set the fear force coefficient  smaller between $5$ and $10$, the success rate of the trained model increases to 70\% to 80\%.

In Figure~\ref{fig:4ts-sc}, the success rate of all methods decreased when the group behaviors from  low separation  and  high cohesion to high separation and low cohesion, but the success rate of the learned model is higher than the other rule-based method. Combining Figures~\ref{fig:4ts-fear} and~\ref{fig:4ts-sc} together, it appears that the fear force coefficient has a greater influence on the success rate than the combination of separation and cohesion coefficients in this particular environment.  

\textbf{Completion Time}.
We study the completion time of successful cases and the time of both successful and failed cases as same as before.

\begin{figure}[th]

\centering
{\includegraphics[width = 0.48\textwidth]{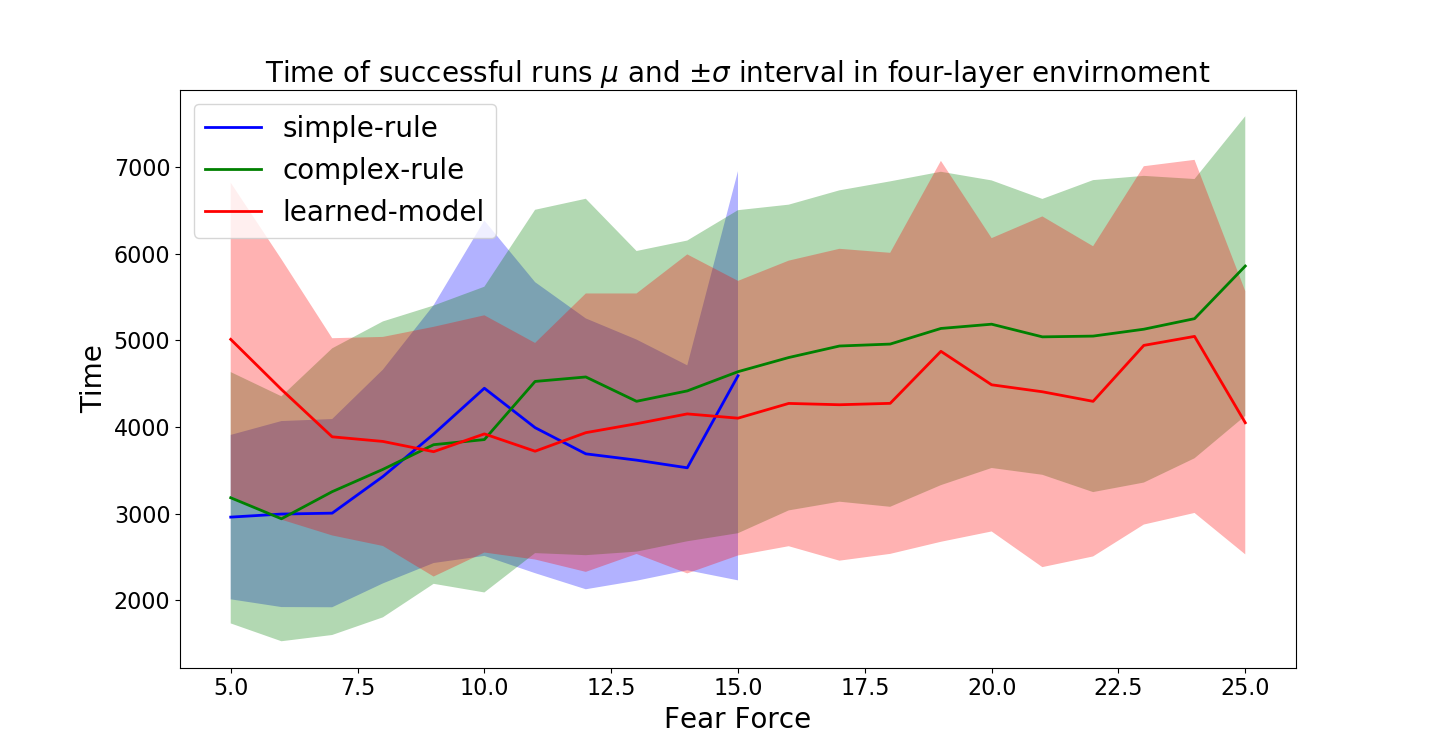}}
{\includegraphics[width = 0.48\textwidth]{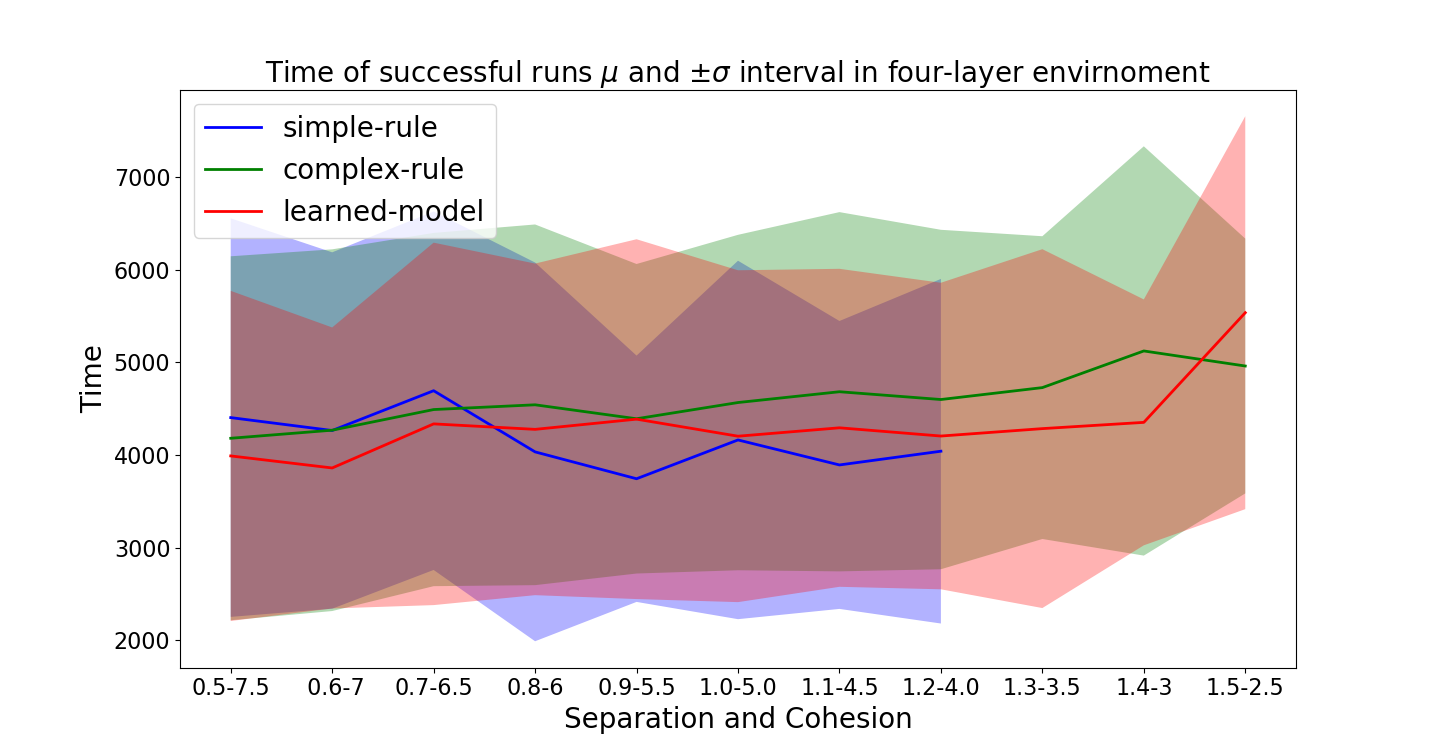}}
\caption{Time of success runs for different levels of group behavior parameters in four-layer environment  }
\label{fig:4ts-stime}
\end{figure}

Figure~\ref{fig:4ts-stime} shows the completion time of successful cases for different levels of group behavior parameters in a four-layer environment. The results show that shepherd with the learned model spends less  time on average than the shepherd with the complex-rule method. The completion time of  the learned model seems  to have  no significant difference 
comparing to that of the simple-rule method. For the standard deviation of average completion time, there is no significant difference among those methods.

\begin{figure}[th]

\centering
{\includegraphics[width = 0.48\textwidth]{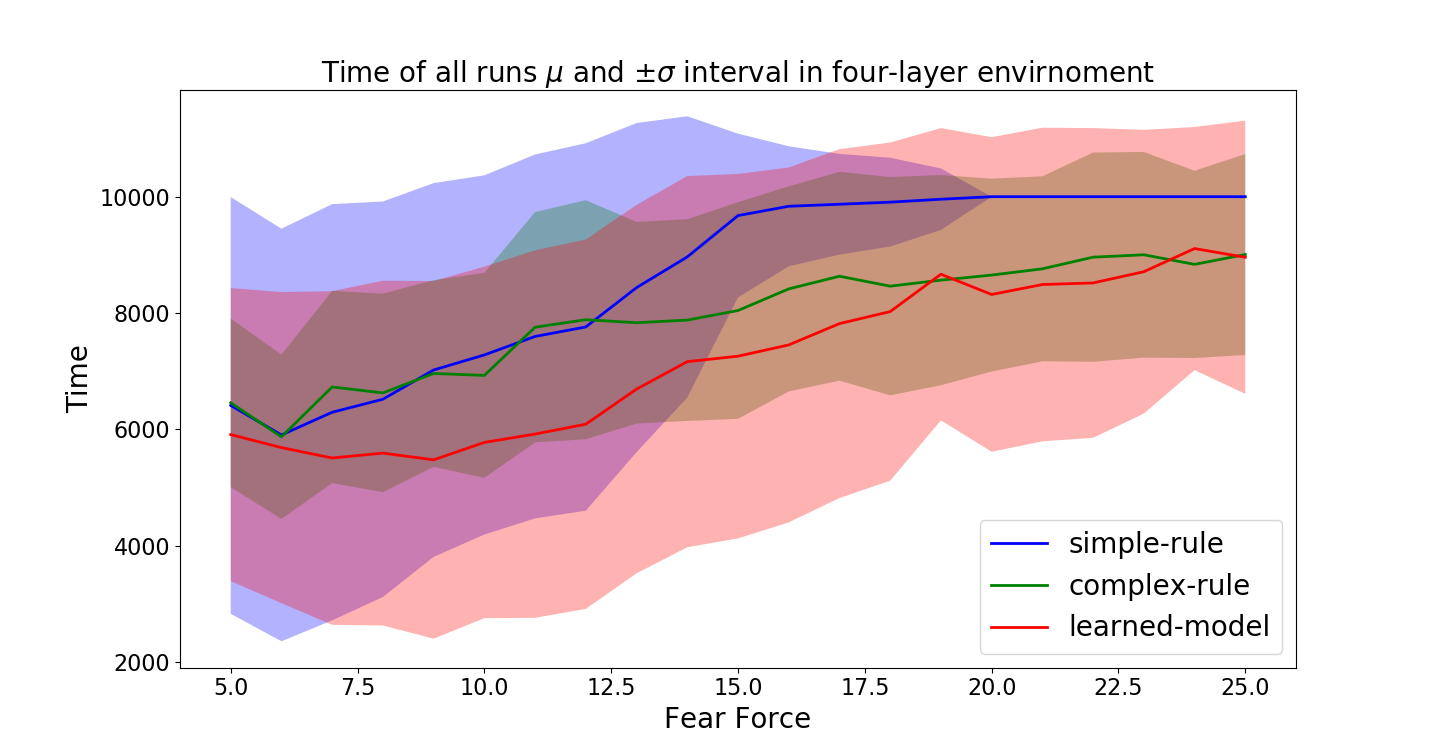}}
{\includegraphics[width = 0.48\textwidth]{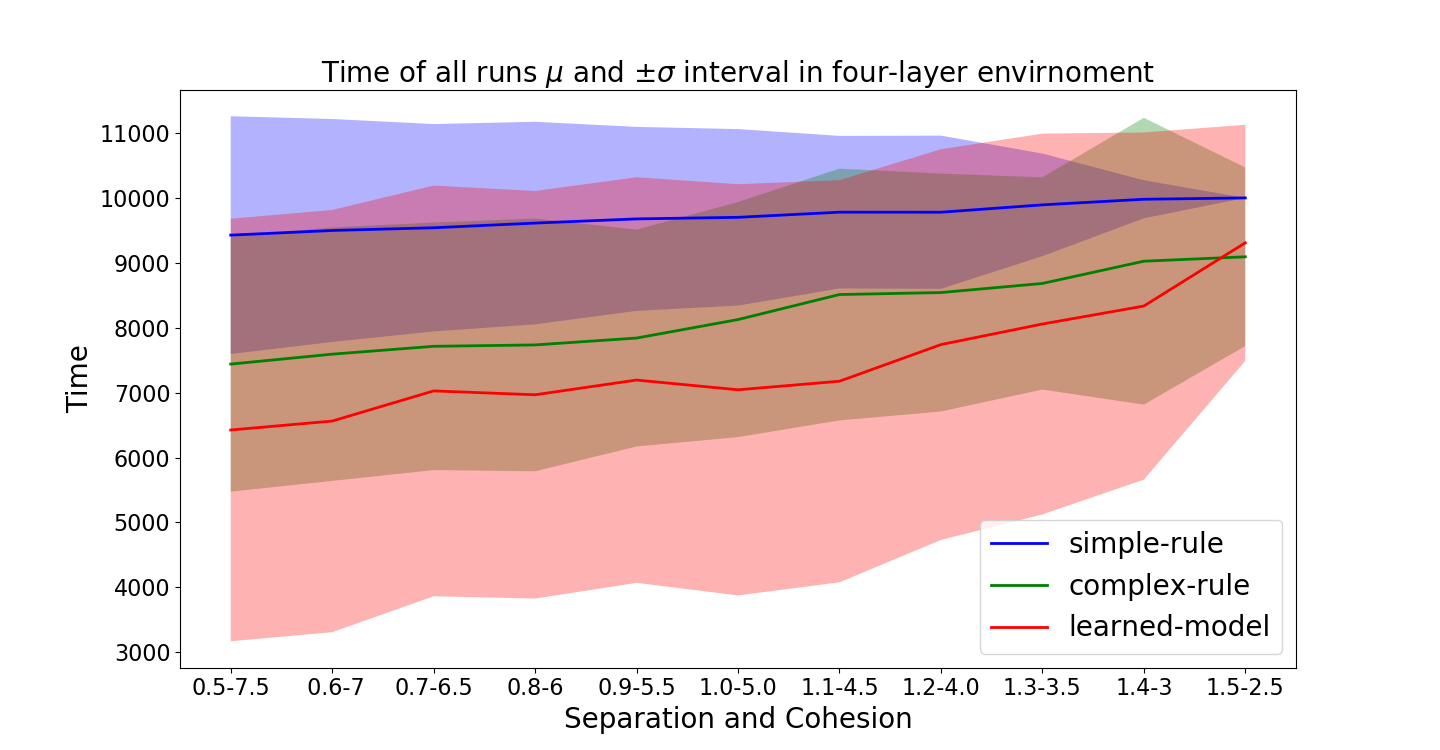}}
\caption{Time of all cases for different levels of group behavior parameters in four-layer environment  }
\label{fig:4ts-atime}
\end{figure}

We analyze the results of successful and failed cases as before.
Figure~\ref{fig:4ts-atime} shows the completion time of all cases using different levels of group behavior parameters in a four-layer environment. 
As given in the figure, a shepherd with the learned model spent less average time than the other two methods due to its higher success rate and shorter herding time.

\begin{figure}[th]
\centering
{\includegraphics[width =0.48\textwidth]{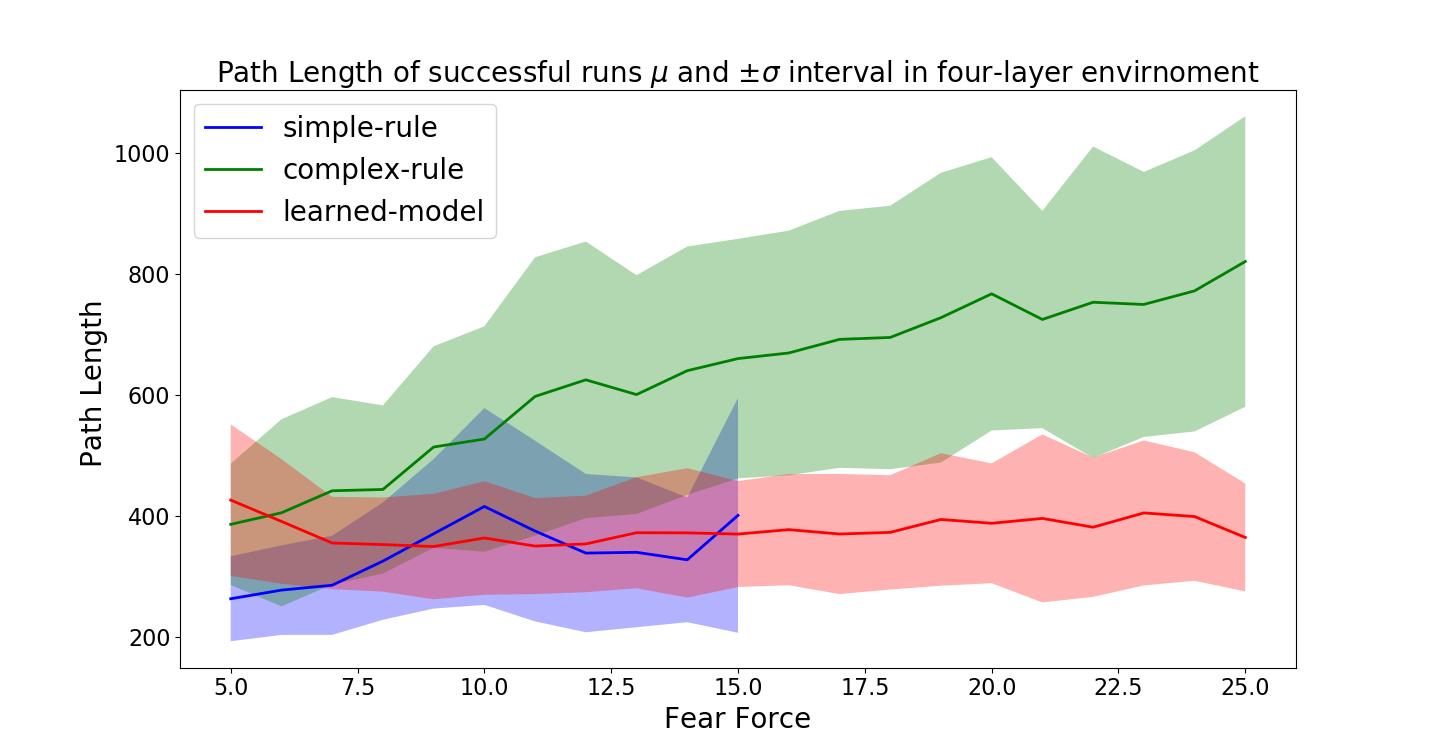}}
{\includegraphics[width = 0.48\textwidth]{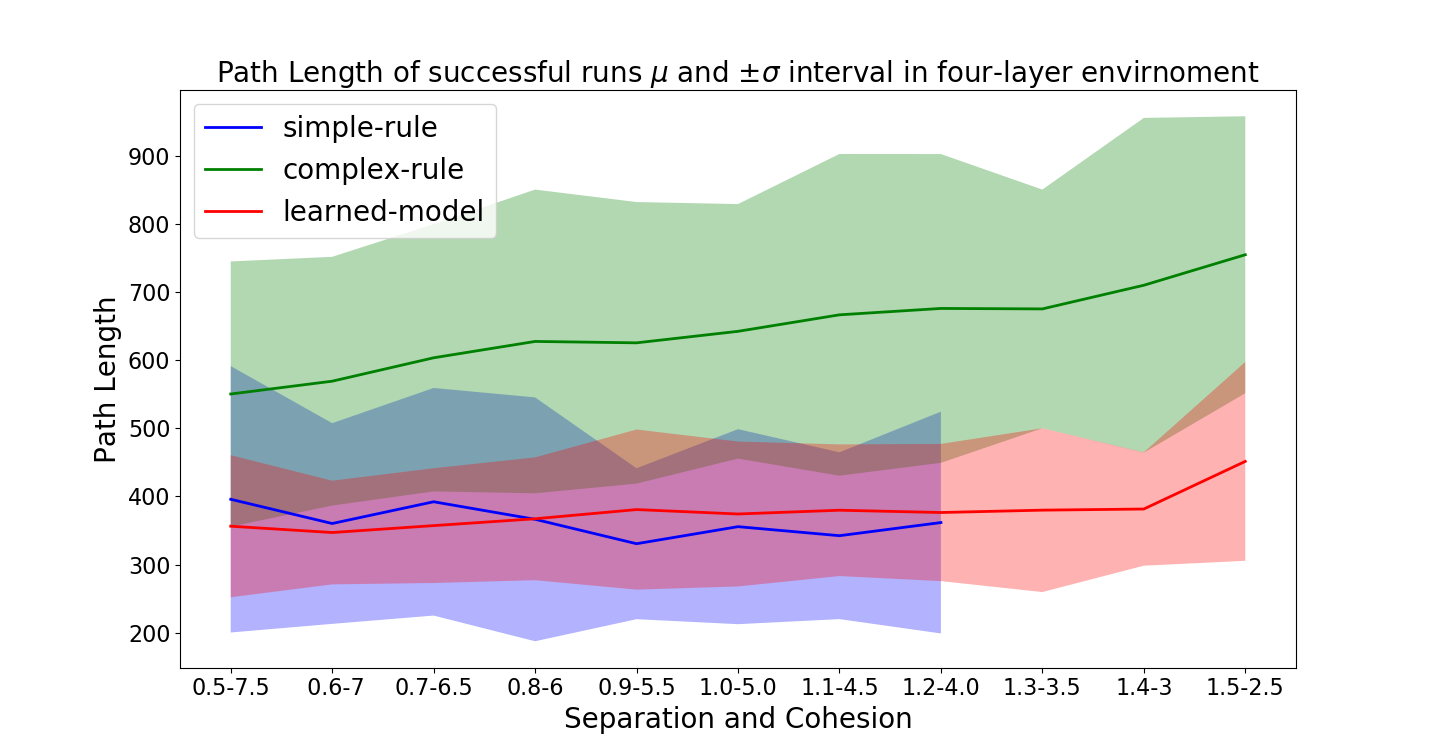}}

\caption{Path length of success runs for different levels of group behavior parameters in four-layer environment.}
\label{fig:4ts-spath}
\end{figure}

\textbf{Path Length}.
Figure~\ref{fig:4ts-spath} shows the path length of successful cases. 
In these figures, the shepherd with the learned model move less than the shepherd controlled by the complex-rule algorithms.
The standard deviation of path length obtained from the learned model is also smaller than the other two methods. 

\section{Conclusion and Future Work}
\label{sec:con}

In this paper, we applied a deep reinforcement learning method to the shepherding problem. We trained the model to herd a group of agents in environments populated with obstacles. 
Our experiments demonstrated that the proposed method is robust to the uncertainties originated from the behavioral model of the group and
environmental model of the obstacles. We showed that the proposed learning-based method has higher  success rate, shorter completion time and path length 
than the traditional rule-based method even in the presence of uncertainties. 

A major limitation of our work is that the proposed method can only handle the group of  2 to 4 agents. The rule based method can handle hundreds of agents.
The second limitation is that our shepherd can only control coherent groups; if the group is severely scattered, the method has poor performance. It will be interesting to see how the proposed work can extend to multiple shepherds to handle both limitations.
 
In the future, we plan to combine rule-based method and learning-based method to provide higher level training of the herding behavior which in turn will provide better shepherding performance. For instance,  the trained model may learn to choose and parameterize a different rule-based method in different environments. In addition, 
we plan to investigate different data representations of the group  other than circle, such as a tight bounding polygon.

\bibliographystyle{IEEEtran}
\bibliography{RL}

\clearpage

\end{document}